\documentclass{article}

\usepackage[preprint]{neurips_2022}

\usepackage[utf8]{inputenc} 
\usepackage[T1]{fontenc}    
\usepackage{hyperref}       
\hypersetup{
    colorlinks=true,
    linkcolor=blue,
    filecolor=magenta,      
    urlcolor=cyan,
    citecolor=blue,
}
\usepackage{url}            
\usepackage{booktabs}       
\usepackage{amsfonts}       
\usepackage{nicefrac}       
\usepackage{microtype}      
\usepackage[dvipsnames]{xcolor}         

\usepackage{aligned-overset}
\usepackage{amsthm}
\usepackage{wrapfig}
\usepackage{diagbox}
\usepackage[toc,page,header]{appendix}
\usepackage{minitoc}

\usepackage[most]{tcolorbox}
\tcbset{
    boxsep=0mm,
    boxrule=0pt,
    colframe=white,
    arc=0mm,
    left=0.5mm,
    right=0.5mm
}
\newcommand{\mb}{\mathbf}

\DeclareMathOperator{\soft}{softmax}
\newcommand{\kro}{%
  \mathbin{\mathop{\otimes}}%
}
\usepackage{xcolor}

\usepackage{thm-restate}
\newtheoremstyle{theoremdd}
  {\topsep}
  {\topsep}
  {\itshape}
  {0pt}
  {\bfseries}
  {. }
  { }
  {\thmname{#1}\thmnumber{ #2}\textnormal{\thmnote{ (#3)}}}
\theoremstyle{theoremdd}

\DeclareMathOperator{\blockdiag}{blockdiag}
\DeclareMathOperator{\diag}{diag}
\DeclareMathOperator{\vect}{vec}

\newcommand{\wQ}{\mb W^{Q}}
\newcommand{\wQT}{\mb W^{Q \top}}
\newcommand{\wK}{\mb W^{K}}
\newcommand{\wKT}{\mb W^{K \top}}
\newcommand{\wV}{\mb W^{V}}
\newcommand{\wVT}{\mb W^{V \top}}
\def\Im{{\bf I}}
\def\Km{{\bf K}}
\def\Am{{\bf A}}

\def\Lm{{\bf L}}
\def\Bm{{\bf B}}
\def\Cm{{\bf C}}
\def\Dm{{\bf D}}
\def\Sm{{\bf S}}
\def\Xm{{\bf X}}
\def\Ym{{\bf Y}}

\def\Mm{{\bf M}}
\def\Tm{{\bf T}}
\def\Wm{{\bf W}}
\def\Zm{{\bf Z}}
\def\Sm{{\bf S}}
\DeclareMathOperator{\tr}{tr}
\newcommand\Exp{\mathbb{E}}

\usepackage{nicematrix}

\newcommand{\norm}[1]{\left\lVert#1\right\rVert}

\newcommand{\extra}[1]{{ #1}}

\usepackage{bm}
\newtheorem{theorem}{Theorem}[section]
\newtheorem{lemma}{Lemma}[section]

\newcommand\blfootnote[1]{%
  \begingroup
  \renewcommand\thefootnote{}\footnote{#1}%
  \addtocounter{footnote}{-1}%
  \endgroup
}

\usepackage{floatrow}

\title{
Signal Propagation in Transformers: Theoretical Perspectives and the Role of Rank Collapse
}

\usepackage{array,multirow,graphicx}

\author{Lorenzo Noci$^{*1}$\\ 
  \texttt{\small lorenzo.noci@inf.ethz.ch} \\
  \And Sotiris Anagnostidis$^{*1}$\\ 
  \texttt{\small sotirios.anagnostidis@inf.ethz.ch} \\ \And Luca Biggio$^{*1,2}$ \\ 
  \texttt{\small luca.biggio@inf.ethz.ch} \\ \And Antonio Orvieto$^{*1}$ \\ 
  \texttt{\small antonio.orvieto@inf.ethz.ch} \\ \And Sidak Pal Singh$^{*1}$ \\ 
  \texttt{\small sidak.singh@inf.ethz.ch} \\ \And Aurelien Lucchi$^{3}$ \\ 
  \texttt{\small aurelien.lucchi@unibas.ch}
}

\date{tests}

\begin{document}

\doparttoc 
\faketableofcontents 

\maketitle

\begin{abstract}
Transformers have achieved remarkable success in several domains, ranging from natural language processing to computer vision. Nevertheless, it has been recently shown that stacking self-attention layers — the distinctive architectural component of Transformers — can result in rank collapse of the tokens’ representations at initialization. The question of if and how rank collapse affects training is still largely unanswered, and its investigation is necessary for a more comprehensive understanding of this architecture. In this work, we shed new light on the causes and the effects of this phenomenon. First, we show that rank collapse of the tokens’ representations hinders training by causing the gradients of the queries and keys to vanish at initialization. Furthermore, we provide a thorough description of the origin of rank collapse and discuss how to prevent it via an appropriate depth-dependent scaling of the residual branches. Finally, our analysis unveils that specific architectural hyperparameters affect the gradients of queries and values differently, leading to disproportionate gradient norms. This suggests an explanation for the widespread use of adaptive methods for Transformers' optimization. 
\end{abstract}

\blfootnote{$^{1}$Dept of Computer Science, ETH Z\"urich, $^{2}$Robotics \& ML, CSEM SA, Alpnach, Switzerland, $^{3}$Department of Mathematics and Computer Science, University of Basel}
\section{Introduction}
Since its first appearance in~\cite{vaswani2017attention}, the Transformer architecture has revolutionized the field of Natural Language Processing (NLP), achieving remarkable success in tasks such as text classification~\citep{yang2019xlnet}, machine translation~\citep{mtlample}, reading comprehension~\citep{brown2020language} and question answering~\citep{raffel2019exploring} among others. Recent efforts have effectively extended its applicability to computer vision~\citep{dosovitskiy2020image} and other domains ~\citep{baevski2020wav2vec, huang2018music, biggio2021neural, polu2022formal}, further popularizing it outside NLP.

The Transformer operates on inputs comprising a sequence of tokens. At its core, it relies on stacked attention layers, which compute a measure of relevance for the whole sequence by assigning token-wise importance weights --- obtained by matrix multiplication of the \textit{queries} and \textit{keys}, and finally normalized with the softmax function. The output of an attention layer is then a linear combination of the importance weights and the so-called \textit{values}. Then, the architecture includes fully-connected sub-layers, residual connections \citep{resnet2016}, and layer normalization (LN), as illustrated in Fig.~\ref{fig:architecture}.

\begin{figure}[t]
\centering
    \includegraphics[width=\textwidth]{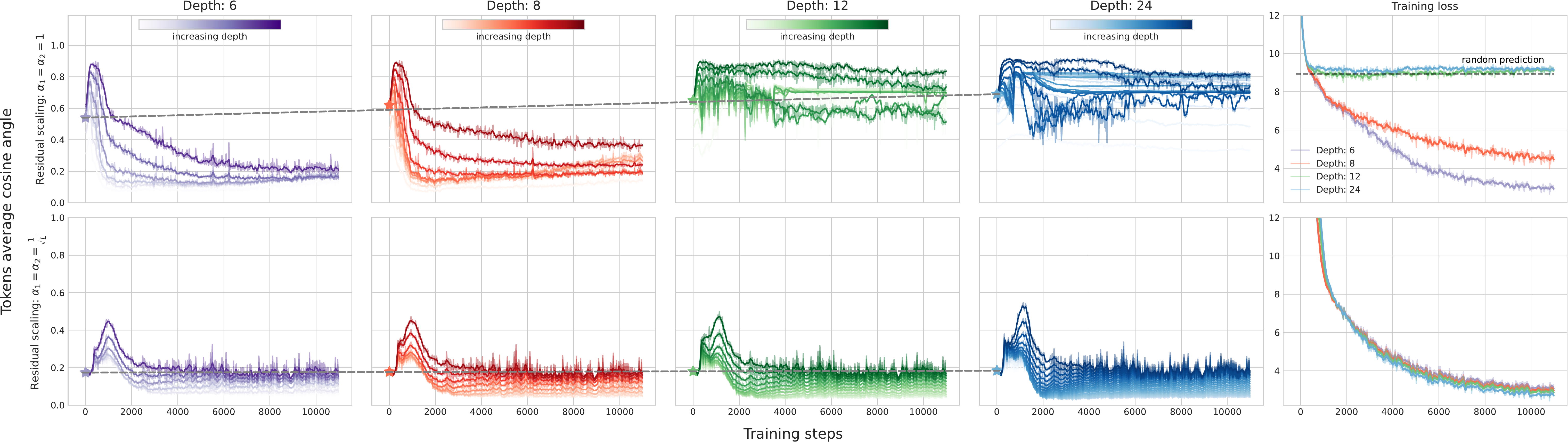}
    \caption{Evolution of the cosine of the angle between tokens for training POST-LN Transformers of increasing depth, with the Adam optimizer, for the IWSLT'14 De-En translation task. Unless adequate residual scaling is used at initialization, increasing depth leads to an increase in the tokens' alignment at initialization, which can inhibit training.}
    \label{fig:adam_postln}
\end{figure}

In the absence of residual connections, \cite{dong2021attention} proved that at initialization the rank of the sequence representation collapses doubly exponentially with depth, and both layer normalization and fully connected layers can only partially alleviate the speed of degeneracy. Under \textit{rank collapse}, the model does not distinguish between representations of different tokens, which are perfectly aligned in feature space at initialization. Crucially, the precise implications of rank collapse in Transformers are not fully understood. 

In this paper, we show that a high alignment of the tokens' representations at initialization --- corresponding to rank collapse in the extreme case of perfect alignment --- affects training by causing vanishingly small gradients of the queries and keys' parameter matrices. This problem severely diminishes the capabilities of the model to learn meaningful attention weights and is further exacerbated in very deep networks, where the rank deficiency --- and hence the vanishing gradient problem of the queries and keys --- affects several layers (see Fig.~\ref{fig:adam_postln}). In order to shed light on this problem, we take inspiration from the flourishing literature on signal propagation in random networks and start our analysis by computing the expected gradients of an attention layer with respect to the queries, keys, and values, which leads to Theorem \ref{thm:vanishing_gradients} on the vanishing gradients for the queries and keys. From here, we pursue two different directions. 

Firstly, we investigate under which conditions rank collapse can be avoided by studying the evolution of the input sequence in a Transformer at initialization. Our theory reveals that a depth-dependent scaling of the residual branches, beyond stabilizing the norm of the activations at initialization, also approximately preserves the cosine of the angle between tokens, and hence also stabilizes the rank of the propagating sequence. We show that this holds even in the infinite-depth limit.

Secondly, we illustrate that there are factors, other than the average tokens' correlation, that affect differently the gradient norm of the queries and keys compared to the values. In particular, the propagating sequence's squared norm has a linear dependence in the values, while a cubic one in the queries and keys, justifying the use of layer normalization. We also highlight a different dependence on the embedding dimension and the length of the input sequence, implying that the gradient norm of a subset of parameters can potentially be of different orders of magnitude, as empirically hinted by previous works \citep{liu2020understanding}. Our analysis brings to light fundamental issues in the signal propagation in Transformers, opening the way for new, well-founded and motivated approaches to improve optimization in these models.

\section{Background}

\paragraph{Transformers.}

\begin{wrapfigure}{r}{0.25\textwidth}
\vspace{-1.5em}
\includegraphics[width=3.5cm]{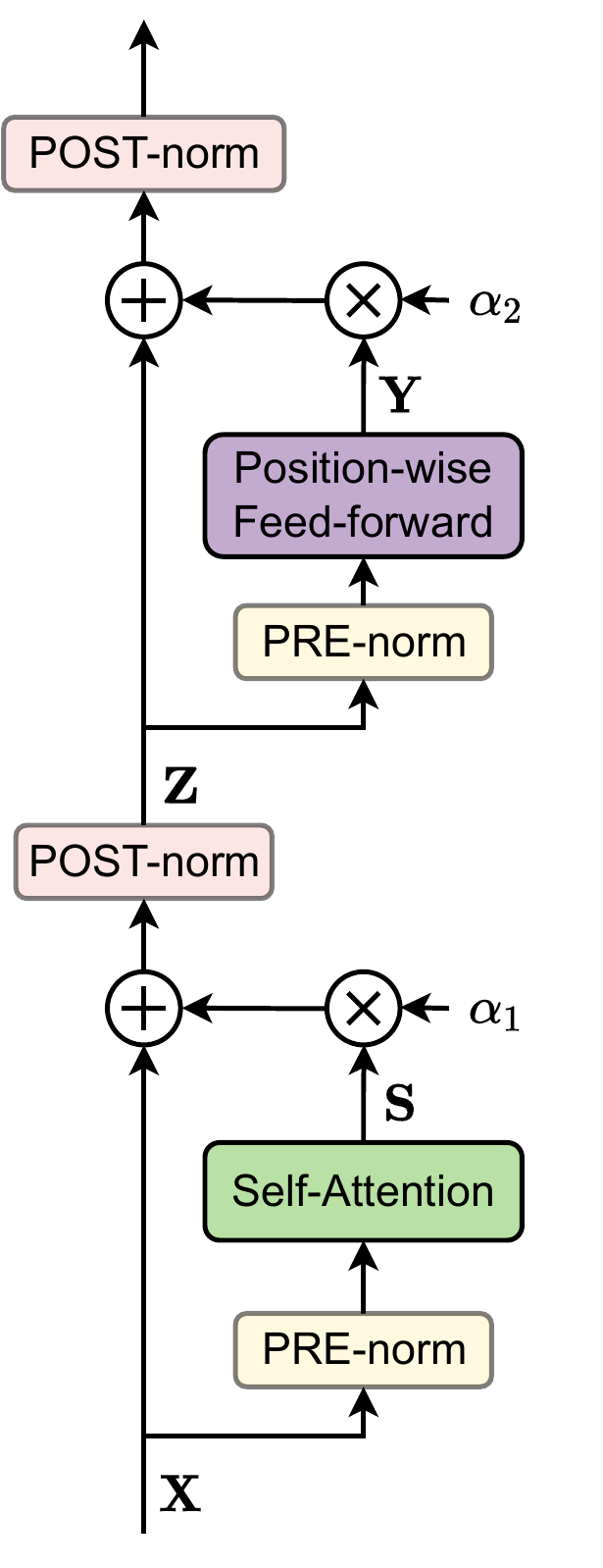}
\vspace{-2em}
\caption{A single Transformer block.\vspace{-2.5em}}
\label{fig:architecture}

\end{wrapfigure}

A Transformer architecture consists of $L$ stacked attention blocks, as show in Fig.~\ref{fig:architecture}. Layer normalization is usually applied token-wise either after the residual connections or to the inputs of the self-attention and position-wise feed-forward sub-layers, leading to the POST-LN~\citep{vaswani2017attention} and PRE-LN~\citep{wang2019learning,xiong2020layer} variants respectively.

Formally, given an input sequence $\Xm \in \mathbb{R}^{n \times d_{v}}$, with $n$ tokens of dimension $d_{v}$, the single-head unmasked scaled dot-product self-attention\footnote{Our analysis also easily generalizes to the case of cross-attention.} is defined as:
\begin{equation}
\label{eq:self_att}
    \Sm^{\ell} := \Am^\ell \Xm^\ell \Wm^{V} , \text{ where } \Am^\ell = \text{softmax}\left( \frac{1}{\sqrt{d_k}}\Xm^{\ell}\Wm^{Q}\left(\Xm^{\ell}\Wm^{K}\right)^\top \right) ,
\end{equation}
where the softmax function is applied independently across each row, and the superscript $\ell$ indexes the $\ell$-th layer.
The matrices $\Wm^Q, \Wm^{K} \in \mathbb{R}^{d_v \times d_k}$ and $\Wm^{V} \in \mathbb{R}^{d_{v} \times d_v}$ are learnable parameters, and each layer is initialized with an independent set of weights. In the literature, the matrices $\Xm^{\ell}\Wm^{Q}, \Xm^{\ell}\Wm^{K}, \Xm^{\ell}\Wm^{V}$ are referred to as queries, keys and values, respectively. The complete Transformer block, in the absence of layer normalization, can be written recursively as:

\begin{align}
    & \Zm^{\ell} = \alpha_1 \Sm^{\ell}+ \Xm^{\ell} \\
    & \Ym^{\ell} = \sigma(\Zm^{\ell} \Wm^{F_1})\Wm^{F_2} \\
    & \Xm^{\ell+1} = \alpha_2 \Ym^\ell +  \Zm^\ell ,
 \end{align}
where the introduced $\alpha_1, \alpha_2$ parameters indicate the strength of the residual block, $\Wm^{F_1}$, $\Wm^{F_2} \in \mathbb{R}^{d_v \times d_v}$ \footnote{In practice, one commonly uses $\Wm^{F_1} \in \mathbb{R}^{d_v \times d_F}$, $\Wm^{F_2} \in \mathbb{R}^{d_F \times d_v}$ where $d_F = \gamma d_v$, with $\gamma \in \{2, 4, 8\}$. Our results then hold up to a constant factor that depends on $\gamma$.} are matrices of learnable parameters; we set $\Xm^0 := \Xm$, and $\sigma: \mathbb{R} \rightarrow \mathbb{R}$ is an activation function. In our case, $\sigma$ is the ReLU function, but we relax this assumption to the linear activation from Section \ref{sec:forward_pass} on.

At initialization, each weight is sampled independently from a distribution with zero-mean and variance $\sigma^2_v = \frac{1}{d_v}$ for the values and feedforward weights\footnote{One should explicitly write the layer dependence $\Wm^{Q,\ell}, \Wm^{K,\ell},\Wm^{V,\ell}, \Wm^{F_1, \ell},  \Wm^{F_2, \ell} $. We at times suppress the $\ell$ index to improve readability. In case $\sigma$ is the ReLU function, we set $\Wm^{F,1}$ to have variance $\frac{2}{d_v}$.}, and $\sigma^2_k = \frac{1}{d_k}$ for the queries and keys. This is the standard ``Xavier''~\citep{glorot2010understanding} or ``He''~\citep{he2015delving} initialization, commonly used in deep learning.  

\paragraph{Rank Collapse in Transformers.}
Interestingly, \cite{dong2021attention} proved that when the residual branches are omitted, the matrix of the tokens' representations $\Xm^\ell$ converges to a rank-1 matrix in which all the representations are the same and equal to a vector $\bm{x}\in \mathbb{R}^{d_v}$, i.e. $\Xm^\ell \to \bm{1}_{n}\bm{x}^\top$, where $\bm{1}_{d_v}$ is the vector with all ones in $\mathbb{R}^{d_v}$. Note that this is a slightly stronger notion of a rank-$1$ matrix, as it implies that all the tokens' representations are both perfectly aligned and have the same norm. Indicating the inner product with the usual bracket notations $\langle \cdot, \cdot \rangle$, and the cosine of the angle between two tokens as $\theta_{k,k'}$, perfect alignment happens when $\langle \Xm^{\ell}_{k}, \Xm^{\ell}_{k'} \rangle = \norm{\Xm^{\ell}_{k}} \norm{\Xm^{\ell}_{k}} \cos \theta_{k,k'}$ with $\cos \theta_{k,k'}=1$ for all $k, k' \in [n]$. Note that perfect alignment together with equal norm between all the tokens implies that all the representations are the same. One of our main contributions is to provide an explanation of how rank collapse affects the gradients of a Transformer at initialization.

\paragraph{Vanishing Gradient Problem.}
Traditionally considered one of the core issues that prevents successful training, the vanishing gradient problem has a long and rich history that dates back to before the popularization of deep learning \citep{hochreiter1991untersuchungen, bengio1994learning}. In its essence, given a loss function  $\mathcal{L}: \mathbb{R}^{n \times d_v} \to \mathbb{R}$, vanishing gradients occur when the norm of the gradient of the loss $\mathcal{L}$ with respect to the parameters of the network $\Wm$ --- which we indicate as $\norm{\frac{\partial \mathcal{L}}{\partial \Wm}}$ --- is too small to provide enough backpropagating signal, thus hindering gradient-based optimization methods. Despite extensive research toward understanding and overcoming the problem in disparate contexts \citep{glorot2010init,he2015delving,hanin2018neural,zhang2019fixup}, a formal explanation of its role in relatively new architectures such as Transformers is largely missing in the literature, with a few exceptions ~\citep{xiong2020layer,wang2022deepnet, huang2020improving}. In our paper (Section \ref{sec:vanishing_gradients}), we show how vanishing gradient occurs in conjunction with the rank collapse issue identified by \cite{dong2021attention}.

\paragraph{Signal Propagation in Random Networks at Initialization.}
After addressing the question on the effects of rank collapse, we take a step back and rigorously analyze its causes by looking at how the properties of the input sequence $\Xm$ are lost/preserved as it propagates through a randomly initialized Transformer. More specifically, we focus on two aspects of the propagating sequence: the expected Frobenius norm $\Exp \norm{\Xm^\ell}^2$ and the expected inner product between different tokens $\Exp \langle\Xm_k, \Xm_k'\rangle$, with $k \neq k'$. The former is linked to a number of studies on the initialization of neural networks at the \emph{edge of chaos}  \citep{poole2016exponential, schoenholz2016deep}, and vanishing/exploding gradients \citep{hanin2018neural}. The latter quantity describes how the geometry of the feature space changes after applying a Transformer block, and is related to the concept of \emph{dynamical isometry} \citep{saxe2013exact}. To understand the evolution of the inner product, we analyze the following measure of correlation \citep{nachum2021johnson, cho2009kernel}: 
\begin{equation}
\label{eq:tokens_corr}
    \rho^\ell_{kk'} := \frac{\Exp\langle\Xm^\ell_{k} , \Xm^\ell_{k'}\rangle}{\sqrt{\Exp\norm{\Xm_k^\ell}^2\Exp\norm{\Xm_{k'}^\ell}^2}}.
\end{equation}
Note that $\rho^\ell_{kk'} = 1$ if and only if the $k$-th and $k'$-th tokens are perfectly aligned ($\cos \theta_{kk'}=1$). We stress that in our case --- differently from the aforementioned works --- instead of analyzing the relationship between two different data points, we study the relationship between tokens of the same sequence. 


\section{Theoretical Results}
\label{sec:theory}

\subsection{Vanishing Gradients for Queries and Keys under Rank Collapse}
\label{sec:vanishing_gradients}
To investigate the problem of vanishing gradients in the attention layers, we make use of the framework of matrix calculus~\citep{magnus2019matrix,singh2021analytic}. In particular, we compare the expected Frobenius norm of the gradient of a self-attention layer with respect to its parameters: $\mathbb{E}\left\|\frac{\partial \Sm^{\ell}}{\partial \Wm}\right\|^2_F$, where here $\Wm$ indicates one of the keys, queries or values weight matrices. 
Due to the well-known difficulty of computing expectations of the softmax \citep{daunizeau2017semi, shekhovtsov2018feed}, throughout this manuscript, we make the simplifying assumption that the softmax output is the uniform distribution at initialization, i.e. the $n \times n$ matrix containing $\frac{1}{n}$ in each entry.
\begin{restatable}[Uniform attention]{ass}{uniformsoftmax}
\label{ass:uniform_softmax}
We assume that $\Am^\ell = \frac{1}{n} \bm{1}_{n\times n}$,
\end{restatable}
where $\bm{1}_{n \times n}$ is the matrix with all entries equal to $1$.
Crucially, in Appendix \ref{app:assumption_unif_soft}, we formally show that \emph{this assumption holds almost surely} in the limit $d_k\to\infty$. There, we also experimentally show that even in the more realistic case where $d_k = d_v \approx 512$, the empirical simulations provide a surprisingly faithful approximation of the theoretical insights presented in this paper.

We define the mean token $\bar{\bm{x}}^{\ell}$ through its components $\bar{\bm{x}}^{\ell}_i = \frac{1}{n}\sum_{k=1}^n \Xm^{\ell}_{ki}$, $i \in [d_v]$. In the following theorem, we compute the expected gradients of an attention layer at initialization, and set the basis for our following analysis. We provide the results only for the queries, as the case for the keys is analogous.

\begin{tcolorbox}
\begin{restatable}[]{lemma}{gradients}
\label{lemma:gradients_queries}
Let $\Xm^{\ell}$ be the representations of the input sequence at the $\ell$-th layer. Under the uniform-attention assumption, we have
    \begin{align}
    \mathbb{E}\left\|\frac{\partial \Sm^{\ell}}{\partial \Wm^{V,\ell}}\right\|^2_F &= d_v n \mathbb{E}\|\bar{\bm{x}}^{\ell}\|^2~\label{eq:jacobian_values};\\ \mathbb{E}\left\|\frac{\partial \Sm^{\ell}}{\partial \Wm^{Q,\ell}} \right\|^2_F &= \frac{\sigma^2_v\sigma^2_k d_v}{n^2}\cdot \Exp \left[ \|\Xm^{\ell}\|^2_F \cdot  \|(\Xm^{\ell})^\top\Xm^{\ell} - n\bar{\bm{x}}^{\ell}(\bar{\bm{x}}^{\ell})^\top\|^2_F\right]~\label{eq:jacobian_queries};\\ 
    \mathbb{E}\left\|\frac{\partial \Sm^{\ell}}{\partial \Xm^{\ell}}\right\|^2_F &\leq \frac{8\sigma^2_q\sigma^2_k\sigma^2_v d_kd_v}{n} \, \cdot \mathbb{E} \norm{(\Xm^{\ell})^\top\Xm^{\ell} - n\bar{\bm{x}}^{\ell}(\bar{\bm{x}}^{\ell})^\top}^2_F + 2d_v^2\sigma^2_v \; .
\end{align}
\end{restatable}
\end{tcolorbox}
We defer the precise study of the scaling of these quantities as a function of $n$ and $d_v,d_k$, to Section~\ref{sec:dep_angle}.
At this stage, it is crucial to note that $\frac{1}{n} (\Xm^{\ell})^\top\Xm^{\ell} - \bar{\bm{x}}^{\ell}(\bar{\bm{x}}^{\ell})^\top$ is the centered empirical covariance matrix of the tokens' representations. It is easy to see that if $\Xm^{\ell}$ is a rank-$1$ matrix, then all the rows of $\Xm^{\ell}$ are proportional to a fixed $d_{v}$-dimensional vector, and the empirical covariance matrix has all zero entries. Introducing a loss function $\mathcal{L}: \mathbb{R}^{n \times d_v} \to \mathbb{R}$, we make the statement on vanishing gradients more formal in the following theorem:
\vspace{15px}
\begin{tcolorbox}
\begin{restatable}[Vanishing gradients under rank collapse]{thm}{vanishinggradients}
\label{thm:vanishing_gradients}
    Suppose that the uniform-attention assumption holds. If additionally $\Xm^{\ell}$ for any $l \in [L]$ has rank-1, and there exists a vector $\bm{x} \in \mathbb{R}^{d}$ such that $\Xm^\ell = \bm{1}_n\bm{x}^T$, then:
  \begin{equation}
      \Exp \norm{\frac{\partial \mathcal{L}}{\partial \Wm^{Q,\ell}}}_F^2 = 0 , \;\;\;\; \Exp \norm{\frac{\partial \mathcal{L}}{\partial \Wm^{K,\ell}}}_F^2 = 0 ,
  \end{equation}
  where the expectation is taken over the weight matrices. This implies that these quantities are vanishing almost surely, due to the non-negativeness of the norm.
\end{restatable}
\end{tcolorbox}
\vspace{15px}

The proof simply relies on expanding the norm of the gradient of the loss with the aid of the chain rule and then bounding it by the product of the norms of each term of the chain. The final result holds with an application of Lemma~\ref{lemma:gradients_queries}, in which the rank-$1$ assumption makes $\mathbb{E}\norm{\frac{\partial \Sm^{\ell}}{\partial \Wm^{Q,\ell}}}$ vanish.
\vspace{15px}
In light of Theorem \ref{thm:vanishing_gradients}, we can conclude that the rank collapse issue originally identified in \cite{dong2021attention} corresponds to an initialization in a region of vanishing gradient signal in the subspace of parameters identified by the queries and keys. How can this affect training? One may argue that if rank collapse does not happen in the very first layer, then the corresponding gradients are non-zero, and the rank of the subsequent layers --- affected by rank collapse --- can be increased with the first few steps of gradient descent. In practice, we show empirically in Fig. \ref{fig:adam_postln} that escaping this pathological landscape is harder in deeper nets.
\vspace{15px}
\subsection{Forward Signal Propagation and the Importance of Scaling the Residual Branches}
\label{sec:forward_pass}
\vspace{10px}
We now turn our attention to the study of the influence of skip connections in transformers. \cite{dong2021attention} showed that simply adding skip connections prevents rank collapse. Somewhat surprisingly, we show that while the claim holds for any finite depth, the average angle between different tokens quickly increases with just a few layers, and as $L\to \infty$ a Transformer can still lose rank unless the residual branches are adequately initialized. As~\cite{dong2021attention} showed that layer normalization does not avoid rank collapse, we omit it in our analysis.
Firstly, we introduce two lemmas on the propagation of inner products (Lemma \ref{lemma:propagation_of_inner_producets}) and the norm (Lemma \ref{thm:forward_pass}) of the tokens' representations. 
\vspace{10px}
\begin{tcolorbox}
\begin{restatable}[Propagation of inner products]{lemma}{propinnprod}
\label{lemma:propagation_of_inner_producets}
 Let $C(\Xm^\ell) = \sum_{k,k'} \langle \Xm_{k}^\ell, \Xm_{k'}^\ell \rangle$ and $\Xm$ the input sequence. Under the Assumption~\ref{ass:uniform_softmax} and if $\sigma$ is the linear activation function, we have that:
 
 \begin{equation}
     \Exp \left[C(\Xm^{L})\right] = (\alpha_2^2 + 1)^{L}(\alpha_1^2 + 1)^{L}C(\Xm)  .
 \end{equation}
 hence, under the depth scaling for the residual block parameters $\alpha_1^2 = \frac{\tilde{\alpha}_1}{L}, \alpha_2^2 = \frac{\tilde{\alpha}_2}{L}$ with $\tilde{\alpha}_1, \tilde{\alpha}_2 \in \mathbb{R}$ independent of $L$, we have that:
 \begin{equation}
      \lim_{L\to \infty} \Exp[C(\Xm^L)] = \text{e}^{\tilde{\alpha}_1 + \tilde{\alpha}_2}C(\Xm).
 \end{equation}
\end{restatable}
\end{tcolorbox}
\vspace{15px}
Note that $C(\Xm^\ell) = n^2 \norm{\bar{\bm{x}}^\ell}^2 $. The lemma on the propagation of the norm is slightly more involved:

\begin{tcolorbox}
\begin{restatable}[Propagation of the norm]{lemma}{forwardpass}
\label{thm:forward_pass}
  Let $\Xm^{L}$ be the representations of the input sequence at the final layer. Under the assumptions of Lemma \ref{lemma:propagation_of_inner_producets}, we have that:
 \begin{equation}
     \Exp \norm{\Xm^{L}}_{F}^2 = n (\alpha_2^2+1)^{L}\alpha_1^2 \sum_{k=0}^{L-1}(\alpha_1^2+1)^k \norm{\bar{\bm{x}}}^2 + (\alpha_2^2+1)^{L} ||\Xm||_F^2  ,
 \end{equation}
 hence, under the depth scaling for the residual block parameters $\alpha_1^2 = \frac{\tilde{\alpha}_1}{L}, \alpha_2^2 = \frac{\tilde{\alpha}_2}{L}$ with $\tilde{\alpha_1}, \tilde{\alpha_2} \in \mathbb{R}$ independent of $L$, we have that:
 \begin{equation}
     \lim_{L\to \infty} \Exp \norm{\Xm^{L}}_{F}^2 = n \text{e}^{\tilde{\alpha}_2}(\text{e}^{\tilde{\alpha}_1} - 1)\norm{\bar{\bm{x}}}^2 + \text{e}^{\tilde{\alpha}_2} ||\Xm||_F^2.
 \end{equation}
\end{restatable}
\end{tcolorbox}
The proof of Lemma \ref{thm:forward_pass} consists in expanding $\Exp \norm{\Xm^{L}}_{F}^2$ according to the defining equations for the Transformer, and simplifying the expression by using iterated expectations $\Exp \norm{\Xm^{L}}_{F}^2 = \mathbb{E}[\Exp [\norm{\Xm^{L}}_{F}^2 | \Xm^{\ell}]]$ to exploit the conditional independence between different layers, and then computing the expectations using the independence assumption on the weights. The expression on the right-hand side will then depend on $ \Xm^{\ell}$ only through its norm $\norm{\Xm^{\ell}}$ and the norm of the mean token $\norm{\bar{\bm{x}}^{\ell}}^2$. Using Lemma \ref{lemma:propagation_of_inner_producets} then allows us to unroll the recursion and get the final result. The complete proof, together with the proof of Lemma \ref{lemma:propagation_of_inner_producets}, can be found in Appendix \ref{app:forward_pass}. 

The previous Lemma provides theoretical justification that scaling the residual branches by setting the alpha parameters to be $\mathcal{O}(1/\sqrt{L})$ allows both the norm of the propagating input and the inner products between different tokens to be approximately preserved. Hence, the information contained in the input is not lost, even in the infinite depth limit.

\begin{figure}[t]
    \centering
    \includegraphics[width=\linewidth]{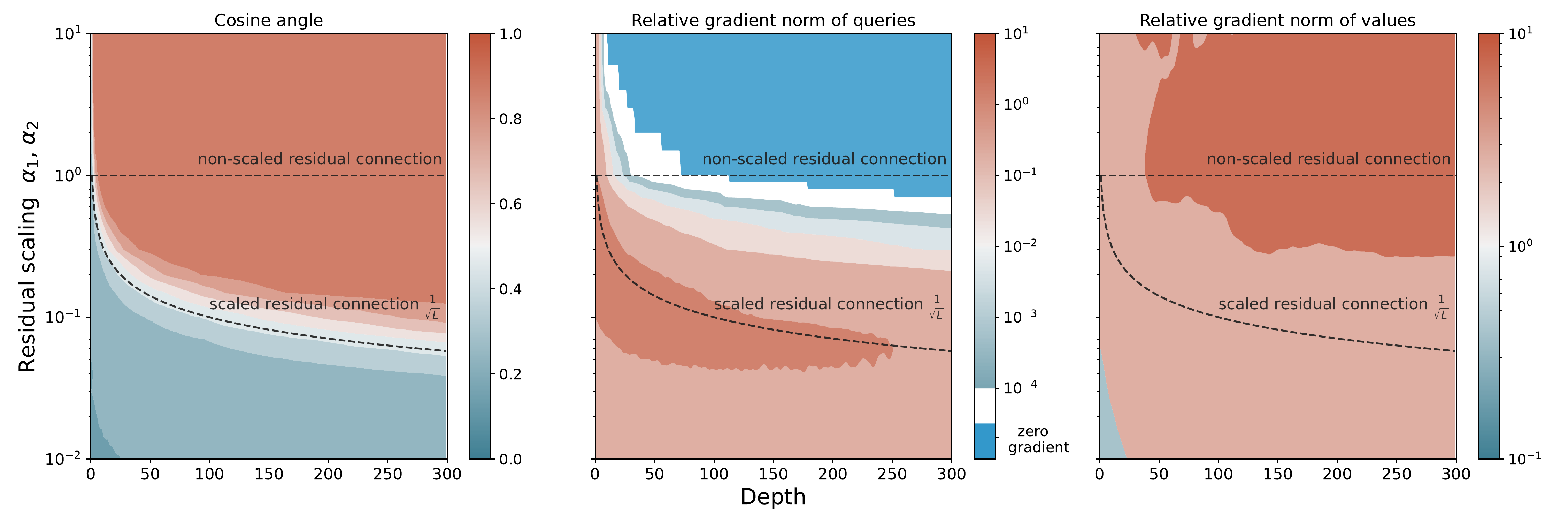}
    \caption{Effect of the residual scaling to the norm of the gradients of the network at initialization with respect to some loss. From left to right: (a) the cosine of the angle between tokens increases with depth. Note how larger values of $\alpha_1, \alpha_2$ imply a faster token alignment with depth (Theorem~\ref{thm:exp_cosine}). Subplots (b) and (c) show the gradients of the queries-keys and values parameters respectively by increasing depth, compared to the corresponding norms of the first layer. Gradients for the queries-keys diminish with depth, while the opposite happens for the values. We use POST-LN to disentangle the effect of the variance of the input.}
    \label{fig:residual_scaling}
\end{figure}

\paragraph{Residual Scaling Preserves Correlations.}
We now prove that without the depth-dependent residual scaling (i.e. with $\alpha_1=\alpha_2=1$) the correlation between the tokens quickly increases, and reaches perfect alignment in the infinite depth limit.
More specifically, our argument shows that in this limit, the correlation between different tokens $\rho^\ell_{k,k'}$ as in Eq.~\eqref{eq:tokens_corr}  converges to 1, implying rank collapse. Furthermore, we show how setting the residual parameters $\alpha_1$ and $\alpha_2$ as dictated by Theorem \ref{thm:forward_pass}, ensures that the correlation measure is dependent on the input in a non-trivial way even at infinite depth. 
To this end, we introduce the average correlation at layer $\ell$: 
\begin{equation}
    \rho^\ell = \frac{1}{n(n-1)}\sum_{k\neq k'}\rho^\ell_{kk'} .
\end{equation}
Note that $\rho^\ell = 1$ if and only if every pair of tokens is perfectly aligned.

We are now ready to formalize the influence of the $1/\sqrt{L}$-scaling on the correlation between tokens' representations by stating Theorem \ref{thm:exp_cosine}.
\begin{tcolorbox}
\begin{restatable}[]{thm}{expectedcosine}
\label{thm:exp_cosine} Let the input tokens have the same norm, i.e. $\norm{\Xm_k} = \norm{\bm{x}} \; \forall k \in [n]$ for some $\bm{x} \in \mathbb{R}^{d_v}$. Under the depth scaling for the residual block parameters $\alpha_1^2 = \frac{\tilde{\alpha}_1}{L}, \alpha_2^2 = \frac{\tilde{\alpha}_2}{L}$ with $\tilde{\alpha}_1, \tilde{\alpha}_2 \in \mathbb{R}$ independent of $L$, we have that: 
  \begin{equation}
      \lim_{L\to \infty}\rho^\ell = \frac{n \text{e}^{\tilde{\alpha}_1}C(\Xm) }{(n-1)[(\text{e}^{\tilde{\alpha}_1} - 1)C(\Xm)  + n\norm{\Xm}_{F}^2]} - \frac{1}{n-1} .
  \end{equation}
 On the other hand, if $\alpha_1, \alpha_2 \neq 0$ are some constants independent of $L$, we have that:
 \begin{equation}
     \lim_{L\to \infty}\rho^\ell = 1.
 \end{equation}

\end{restatable}
\end{tcolorbox}
The proof consists in noting that due to the symmetry of the problem at initialization, for a fixed layer the expected norm of each token is the same. Hence, by our definition of $\rho^\ell_{kk'}$, we can write $\Exp\langle\Xm^\ell_k, \Xm^\ell_{k'}\rangle = \rho^\ell_{kk'} \Exp\norm{\bm{x}^\ell}^2$. By summing over the $k,k'$ indexes, the resulting equation will depend on $\Exp[C(\Xm^\ell)]$ and $\Exp \norm{\Xm^\ell}^2$, which can be expanded using Lemma \ref{lemma:propagation_of_inner_producets} and \ref{thm:forward_pass} respectively. The result is then given by solving for $\rho^\ell$. 

\begin{wrapfigure}{r}{0.4\textwidth}
\vspace{-5px}
\centering
\includegraphics[scale = 0.38]{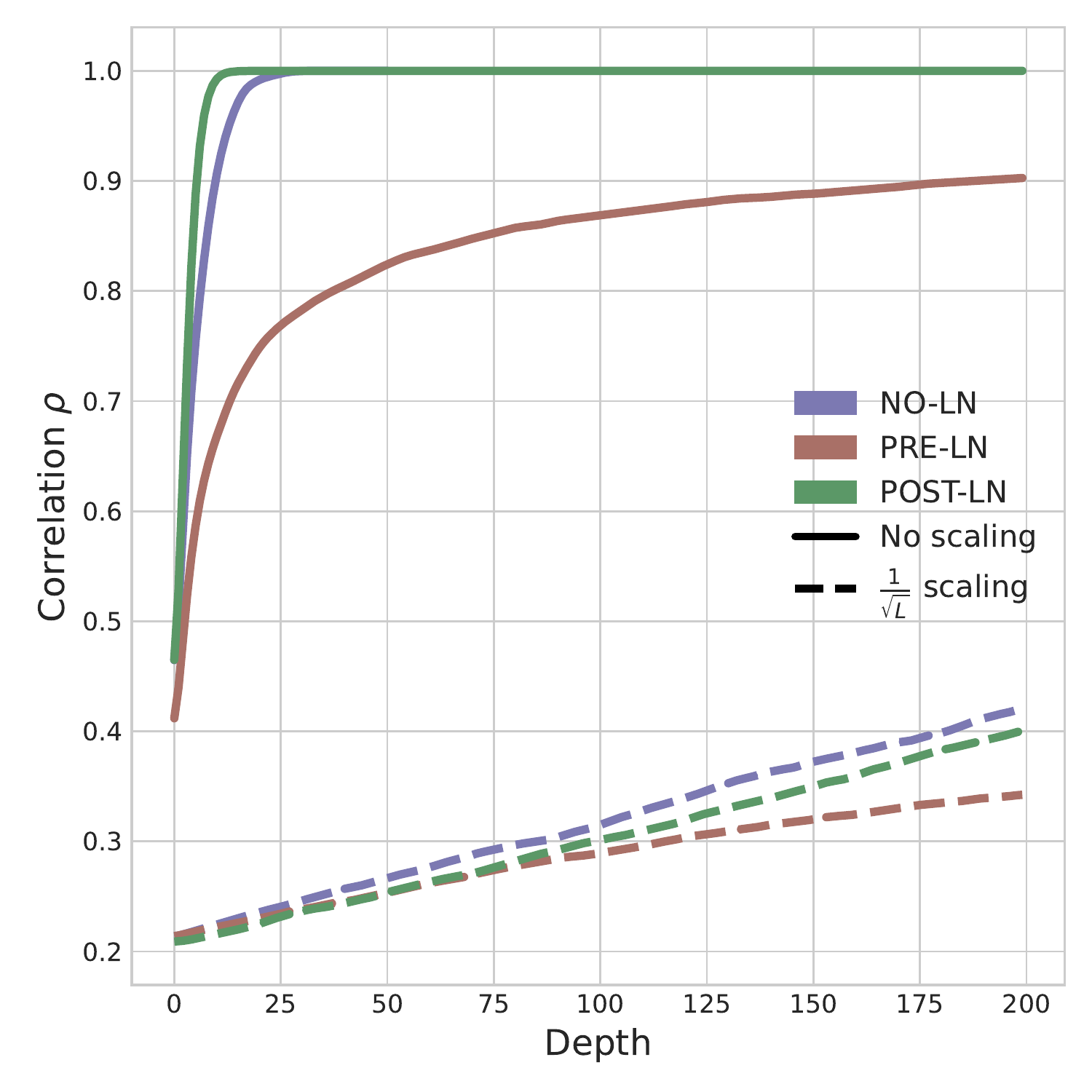}
\caption{Evolution of Correlation in Transformers with (dashed lines) and without (solid lines) $1/\sqrt{L}$-scaling for PRE-LN, POST-LN and without layer normalization (No-LN).}
\label{fig:corr}
\vspace{-5px}
\end{wrapfigure}

\vspace{10px}
Note that under the $1/\sqrt{L}$-scaling, the correlation term is one if and only if $C(\Xm) = n\norm{\Xm}^2$, which holds in the degenerate case where all the input tokens are perfectly aligned. In Appendix \ref{app:res_scaling_proofs}, we give precise formulas for the expected correlations at any depth, showing that $\rho^\ell$ reaches values close to one even for relatively shallow networks when the $1/\sqrt{L}$-scaling is not adopted (see also Fig. \ref{fig:residual_scaling}~(left)). Additionally, in Fig.~\ref{fig:corr}, we empirically show that in the presence of the $1/\sqrt{L}$-scaling, layer normalization (either PRE or POST) does not significantly affect the evolution of the correlations. On the other hand, without the residual scaling, PRE-LN seems to alleviate the rate of increase of $\rho^\ell_{kk'}$. It is intriguing that most deep Transformer models use this configuration~\citep{brown2020language}. We provide more extensive empirical results in Appendix~\ref{app:more_experiments}.

\vspace{10px}
Note that the $1/\sqrt{L}$ scaling for the residual branches has been previously studied in the context of stabilization of residual networks (see Section \ref{sec:related_work}), here we extend these results to Transformers and provide new insights on its role in the context of rank preservation.
Finally, note that by setting $\tilde{\alpha}_1, \tilde{\alpha}_2 = 0$, we recover the so called "ReZero" initialization \citep{bachlechner2021rezero}. In this context, the $1/\sqrt{L}$ scaling extends this framework as it allows for wider range of values for $\tilde{\alpha}_1, \tilde{\alpha}_2$ while still guaranteeing stability. 
We mention here that extending these results from the linear activation to the ReLU case is known to be a hard problem, due to the technical difficulty of propagating the inner products across ReLU layers that are shared among the tokens (this is the case in the position-wise feed-forward layers in Transformers). Exact formulas can be found only in the case of one ReLU layer with Gaussian inputs in \cite{cho2009kernel}. 

\subsection{Dependence on the Angle between Tokens and the Input Norm}
\label{sec:dep_angle}
In this section, we drop the superscript $\ell$ as it is obvious from context and assume for simplicity that $d_k = d_v$. To gain a better intuition on the factors that affect the gradients and provide additional insights, we study the case in which every pair of distinct tokens are  zero-mean Gaussian random variables, correlated in the same way, i.e $\rho^\ell_{ii'} = \rho$ for $i \neq i'$ or more precisely
\begin{equation*}
    \Exp\left[\Xm_{i,j}\Xm_{i',j'}\right] = \begin{cases}
    0 & j\ne j' \ \ \text{(independent dimensions)}\\
    \sigma^2_x & i=i', j=j'\\
    \rho\sigma^2_x & i\ne i', j=j'
    \end{cases}.
\end{equation*}
To see that this equation satisfies our definition of the correlation metric, note that $\Exp[\norm{\Xm_i}^2] = d\sigma_x^2$ and $\Exp\langle \Xm_{i}, \Xm_{i'} \rangle = d\sigma_x^2 \rho$, for $i \neq i'$.
Then, the expected norm of the gradients for the values (Eq.~\eqref{eq:jacobian_values}) simplifies to
\begin{equation}
     \Exp \norm{\frac{\partial \Sm}{\partial \wV}}_F^2 = \sigma_x^2 d^2 \left(1 + \rho (n-1)  \right).
\label{eq:grad_V}
\end{equation}

By making the additional assumption that the norm and the correlation propagate independently, the respective norm for the queries --- and symmetrically the keys --- (Eq.~\eqref{eq:jacobian_queries}) reduces to:
\begin{equation}
    \Exp \left\|\frac{\partial \Sm}{\partial \wQ}\right\|_F^2 = \sigma_x^6 \frac{(n-1)}{n} (1 - \rho)^2 d (n + d).
\label{eq:grad_Q}
\end{equation}
In Appendix~\ref{sec:constan_cosine_analysis} we provide a rigorous proof, that relies on Isserlis theorem~\citep{isserlis1918formula} to compute higher-order moments. The above expressions reveal the different dependencies on four main actors, that we inspect separately here. The gradients of the queries depend via a cubic function on the \emph{variance of the input, $\sigma_x^2$}, compared to a linear for the values. This provides an additional interpretation of the successful use of layer normalization, as in~\cite{xiong2020layer}, either in the POST-LN or PRE-LN format, that standardizes the input variance $\sigma_x^2$ to the value $1$.

Next, we emphasize the dependence on the \emph{correlation between the tokens}, also illustrated in Fig.~\ref{fig:residual_scaling}. Importantly, note how the queries/keys have opposite monotonic functional dependence with respect to $\rho$ compared to the values. As revealed by Theorem~\ref{thm:exp_cosine} and Fig.~\ref{fig:residual_scaling}~(center), inappropriate scaling of the residual branches can already lead to this phenomenon even in a relatively shallow network. 

Finally, Eq.~\eqref{eq:grad_V} and~\eqref{eq:grad_Q} reveal a different scaling in terms of the \emph{embedding size $d$} and the \emph{sequence length $n$} due to the self-attention operation itself. We hope that the identification of the different dependencies in the gradients of the parameters will inspire a new line of works aimed at solving some of the difficulties in training Transformers.

\subsection{Are Adaptive Methods really needed for training Transformers?}
\label{sec:adaptive}

\begin{wrapfigure}{r}{0.50\textwidth}
\vspace{-15px}
\centering
\includegraphics[width=6.5cm]{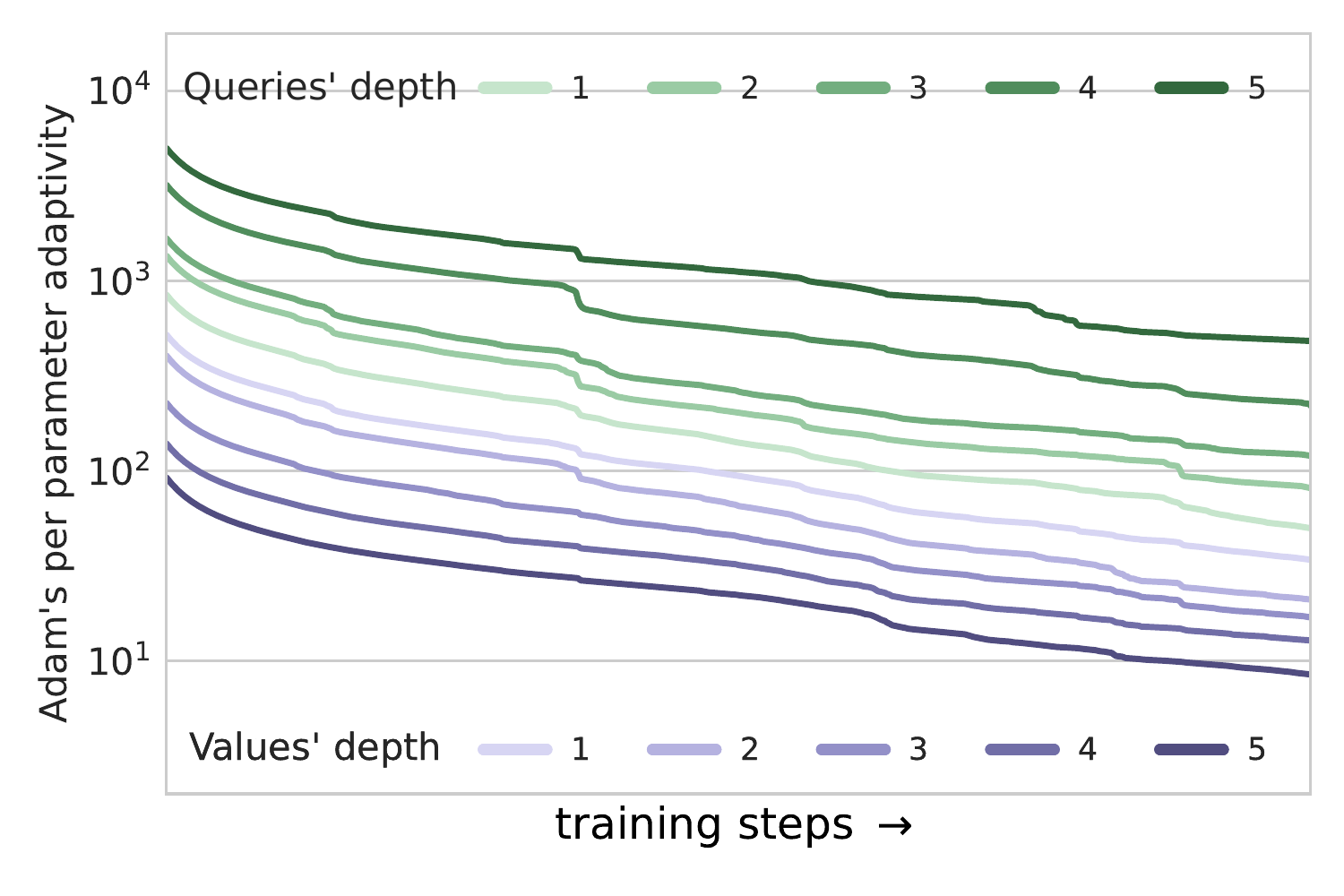}
\caption{Adaptive learning rates computed by Adam in Transformers.}
\label{fig:adam-adap}
\vspace{-10px}
\end{wrapfigure} 

The existence of the discrepancy in the magnitude of the gradients with respect to the weights $\Wm^{Q}, \Wm^{K}$ and $\Wm^{V}$, might explain the success of adaptive optimization algorithms, as illustrated in Fig.~\ref{fig:adam-adap}, where we plot the effective learning rate computed by Adam~\citep{kingma2014adam} in a toy encoder task (more details in Appendix~\ref{app:experimental_setup}). Hence, we embark on a preliminary exploration to train a Transformer architecture with SGD with the intent of matching Adam's performance. 
Based on our theory, we propose a simple architectural modification, an inverse temperature scaling $\tau \in \mathbb{R}$ inside the softmax:
\begin{equation}
    \Sm^{\ell} := \text{softmax}\left( \frac{\tau}{\sqrt{d_k}}\Xm^{\ell}\Wm^{Q}\left(\Xm^{\ell}\Wm^{K}\right)^\top \right) \Xm^\ell \Wm^{V}.
\end{equation}
A direct consequence of our analysis is that $\tau$ allows controlling the magnitude of the gradients for the queries and keys' parameters.

\begin{figure}
\CenterFloatBoxes
\begin{floatrow}
\ffigbox
  {\includegraphics[width=0.5\textwidth]{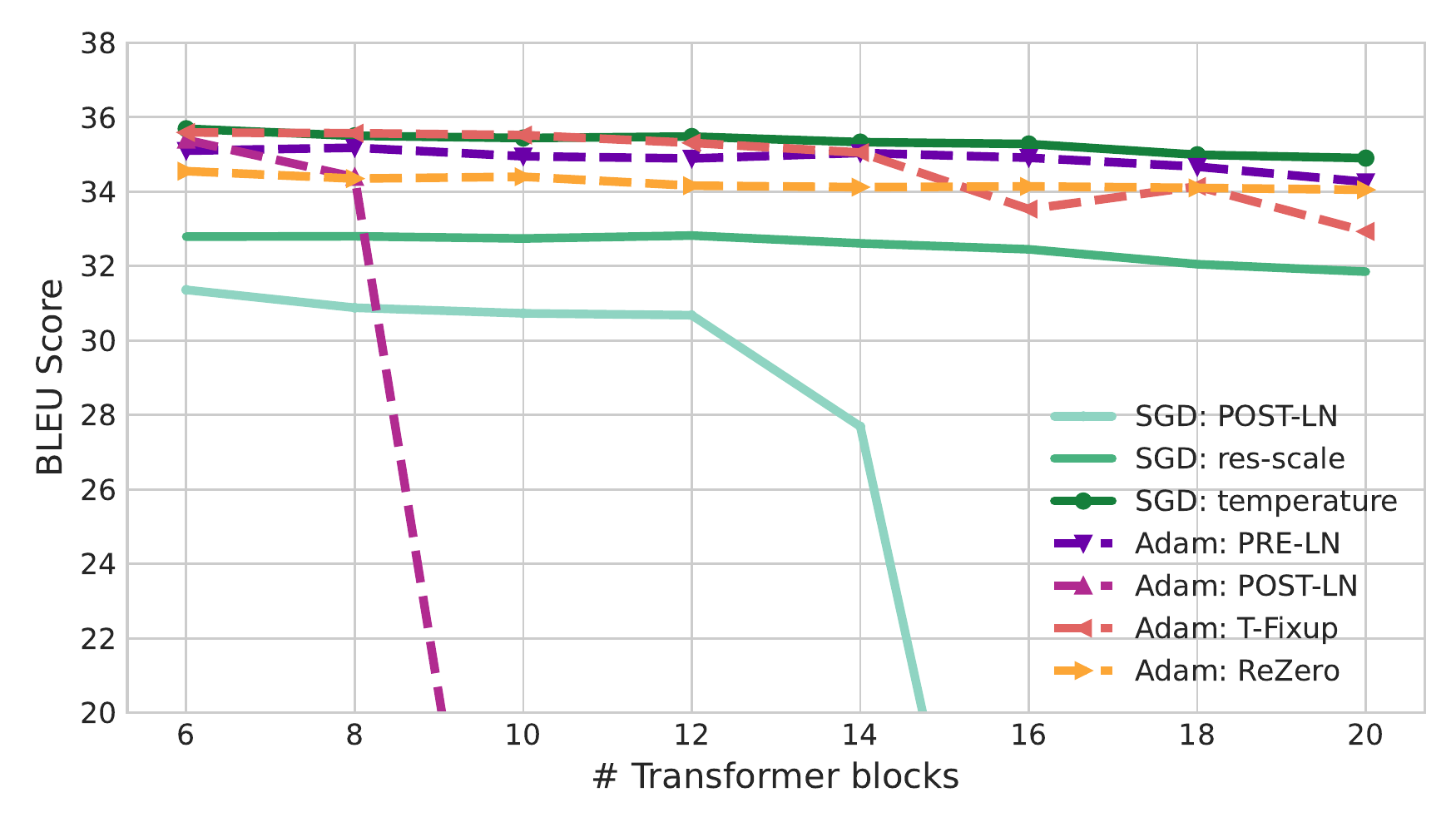}}
  {\caption{\small{BLEU scores by increasing the number of transformers blocks. `X' Transformer blocks implies in total `X' encoder self-attention, `X' decoder self-attention and `X' decoder cross-attention layers.}}\label{fig:translation_fig}}
\killfloatstyle
\ttabbox
   {
   \small
   \renewcommand{\arraystretch}{1.2}
   \begin{tabular}[b]{lc}\hline
     Method (6L-Encoder / 6L-Decoder) & BLEU $\uparrow$ \\ \hline
     SGD POST-LN & 31.36\\
     SGD res-scale & 32.79\\
     SGD temperature & \textbf{35.69} \\ 
     \hline
     Adam POST-LN~\citep{vaswani2017attention} & 35.39\\
     Adam PRE-LN~\citep{vaswani2017attention} & 35.10 \\
     ReZero~\citep{bachlechner2021rezero} & 34.55\\
     T-Fixup~\cite{zhang2019fixup} & 35.59\\
     \hline
     \end{tabular}
  }
  {\caption{\small{BLEU scores for the IWSLT14 German-to-English translation task. \textit{SGD res-scale} refers to the training of SGD without layer normalization and initialization of the residual scaling $a_1 = a_2 = \frac{1}{\sqrt{L}}$. \textit{SGD temperature} additionally employs an inverse temperature inside the softmax.}\label{tab:translation_results}}}
\end{floatrow}
\end{figure}

We evaluate our proposal, consisting of residual scaling and the aforementioned inverse temperature parameters, on the widely used IWSLT14 German-to-English (De-En) benchmark translation task. All details regarding the experimental setup and the choice of inverse temperature used are provided in Appendix~\ref{app:experimental_setup}. We train a Transformer encoder-decoder of varying depth with SGD, after removing all normalization layers and adequately initializing the residual connections. For our training with SGD, we avoid using any learning rate warm-up, as commonly done for Adam, and instead use a step-scheduler to decrease the learning rate at 40\% and 80\% of training. We compare against the following methods that make use of Adam; POST-LN and PRE-LN refer to the aforementioned alternatives to apply layer normalization. We also compare against other successful techniques that rely on specific initializations to avoid layer normalization, such as ReZero~\citep{bachlechner2021rezero} and T-Fixup~\citep{zhang2019fixup}. We report  the average BLEU score~\citep{papineni2002bleu} across 5 runs in Fig.~\ref{fig:translation_fig} and Table~\ref{tab:translation_results}.

Our proposed method considerably improves training with SGD, keeping up and in some cases surpassing any results achieved by the Adam optimizer. We are also able to train deeper networks without the use of layer normalization. We leave for future work to further investigate modifications or alternatives to the self-attention operation.

\section{Related Work}
\label{sec:related_work}
Our work builds upon the rich literature on forward and backward signal
propagation in random neural networks \citep{poole2016exponential,schoenholz2017deepinformationpropagation,pmlr-v80-xiao18a,NIPS2017_d9fc0cdb,orvieto2021vanishing, noci2021precise, zavatone2021exact}.
The $1/\sqrt{L}$ scaling scheme has been investigated in the literature for the stabilization of residual networks \citep{hanin2018start, arpit2019initialize, allen2019convergence, hayou2021stable}.

Our work draws inspiration from a series of recent works studying the rank of the representations of random feed-forward neural networks at initialization \citep{jonas, daneshmand2020batch}. In the context of Transformers, \cite{dong2021attention} has recently identified the rank collapse issue object of study of the present work. Thanks to our analysis of the backward pass, we are able to demonstrate that rank collapse in Transformer architectures leads to vanishingly small gradients of queries and keys, thereby preventing effective training and allowing us to complete the analysis of \citep{dong2021attention}.

Among the architectural components in Transformers, layer normalization is, arguably, one of the most important -- and debated -- ones \citep{chen2018best,wang2019learning,nguyen2010estimating,xiong2020layer}. In the original architecture \citep{vaswani2017attention}, layer normalization is used to stabilize the forward pass by reducing the variance of the inputs to the following sublayer. Our analysis of the forward pass shows that its inclusion is not strictly necessary for the purpose of controlling the norm of the representations. For a theoretical analysis of signal propagation in the presence of layer norm, we refer the reader to \cite{xiong2020layer}.

Additionally, our theoretical study of the backward pass provides a rigorous explanation of the empirically observed discrepancy between the magnitude of the gradients of the queries and the values, which \cite{liu2020understanding} hypothesize to be one of the causes of the success of adaptive methods in training Transformers \citep{liuadam,zhang2020adaptive,huang2020improving}.

Finally, properly rescaled residual connections have been found to be beneficial for training Transformers by a number of recent research works \citep{zhang2019fixup, bachlechner2021rezero, wang2022deepnet}. However, none of these studies characterize the impact of skip connections on rank propagation, while our analysis suggests a theoretically-grounded way to stabilize it.

\section{Conclusions and Future Work}
In this paper, we showed how, at initialization, rank collapse and more generally high correlation in the tokens, causes vanishing gradients of the queries and keys of a Transformer architecture. While residual connections help mitigate rank collapse at finite depth, we showed that they alone cannot prevent high alignments of the tokens' representations --- unless properly scaled by a $1/\sqrt{L}$-factor. Finally, we have also discovered counter-intuitive dependencies on the variance of the input, embedding size, and sequence length, potentially causing large differences between the gradients of queries/keys compared to the values' parameters. Hence, we conclude that one of the strengths of Transformers lies in their carefully designed architecture together with an adequate initialization. 
Finally, we gave preliminary evidence that one of the factors contributing to the higher efficacy of Adam compared to SGD in training Transformers arises from the disproportionate magnitude of gradients as postulated by our theory.
Nonetheless, other factors might further accentuate the difference between these two algorithms during training, leaving the door open for further research regarding the benefits of adaptive optimization methods with Transformers.

\section*{Acknowledgements}

We thank our colleague Jonas Kohler who provided insights and expertise that greatly assisted the research.

\bibliographystyle{plainnat}
\bibliography{main}

\newpage

\appendix
\addcontentsline{toc}{section}{Appendix}
\part{Appendix} 
\parttoc 

\section{Proof of Theorems}
Recall the defining equations of a Transformer:
\begin{align*}
    & \Zm^{\ell} = \alpha_1 \Sm^{\ell}+ \Xm^{\ell} \\
    & \Ym^{\ell} = \sigma(\Zm^{\ell} \Wm^{F,1})\Wm^{F,2} \\
    & \Xm^{\ell+1} = \alpha_2 \Ym^{\ell} +  \Zm^{\ell} ,
\end{align*}
where the self attention layers are defined as follows:
\begin{equation*}
    \Sm^{\ell} := \Am^\ell \Xm^\ell \Wm^{V} , \text{ where } \Am^\ell = \text{softmax}\left( \frac{1}{\sqrt{d_k}}\Xm^{\ell}\Wm^{Q}\left(\Xm^{\ell}\Wm^{K}\right)^\top \right).
\end{equation*}
We remark that when it is clear from the context, we suppress the $\ell$ index to improve readability.

\textbf{Initialization}: Recall that we initialize our weights with the so called ``Xavier''~\citep{glorot2010understanding} or ``He''~\citep{he2015delving} initialization: each weight is sampled independently from a distribution with zero-mean and variance $\sigma^2_v = \frac{1}{d_v}$ for the values and feedforward weights and $\sigma^2_k = \frac{1}{d_k}$ for the queries and keys. 

\textbf{Kronecker Delta}: we introduce the Kronecker Delta notation 
\begin{equation*}
    \delta_{ab} = 
    \begin{cases}
        1 & a=b \\
        0 & a\neq b
    \end{cases}
\end{equation*}
and, similarly: $\delta_{a\neq b} = 1 - \delta_{ab}$.

\textbf{Notation}: in this section, we also adopt the following shorthand notation for the argument of the softmax $\Mm:=\frac{\Xm\wQ{\wK}^\top \Xm^\top}{\sqrt{d_k}}$. We will first compute the gradients with respect to values, queries and input (note that the gradients of the keys have the form as the queries, hence we omit the derivation). Recall that for a matrix $\Xm$, we use $\Xm_k$ to indicate its $k$-th row. Finally, we indicate with $\bm{1}_{n\times n}$ the $n \times n$ matrix with all ones, with $\bm{1}_n$ the columns vector with all ones, and with $\Im_n$ the $n$-dimensional identity matrix. 

\subsection{Backward Pass: Proofs of Lemma \ref{lemma:gradients_queries} and Theorem \ref{thm:vanishing_gradients}}
In this section, we now look at the proofs of Lemma \ref{lemma:gradients_queries} and Theorem \ref{thm:vanishing_gradients}. We will first introduce our notation for the gradients, as well as some useful properties of the Kronecker product.

\subsubsection{Preliminaries}
For the gradients, we avoid directly working with tensors by vectorizing the matrices in a \textit{row-wise} fashion ($\vect_r$) and arranging the Jacobian in the numerator layout. More formally,

$$\dfrac{\partial \mb Y}{\partial \mb X} := \dfrac{\partial \vect_r(\mb Y)}{\partial \vect_r(\mb X)^\top}\,.$$

Alongside this, we use the following rule ($\kro$ is the Kronecker product):
\begin{align}
\label{eq:matrix-derivative}
 \frac{\partial \mb A \mb W \mb B}{\partial \mb W} = \mb A \kro \; \mb B^\top\,.
\end{align}

For the proof of this rule, we refer to \cite{singh2021analytic}, and to \cite{magnus2019matrix} for a complete introduction to matrix calculus. 
We will also use the following well-known properties of the Kronecker product.
\begin{tcolorbox}
\begin{lemma}
\label{lemma:prop_kro}
Given the matrices $\Am \in \mathbb{R}^{n \times m}$, $\Bm \in \mathbb{R}^{p \times q}$, $\Cm \in \mathbb{R}^{m \times r}$, $\Dm \in \mathbb{R}^{q \times s}$, then the following holds:
\begin{equation}
    \label{eq:trace_kro}
    \tr(\Am \kro \Bm) = \tr(\Am)\tr(\Bm) ,
\end{equation}
and 
\begin{equation}
    \label{eq:prod_kro}
    (\Am \kro \Bm)(\Cm \kro \Dm) = (\Am \Cm)\kro(\Bm \Dm).
\end{equation}
\end{lemma}
\end{tcolorbox}

\subsubsection{Proof of Lemma~\ref{lemma:gradients_queries}}
In Lemma \ref{lemma:grads_SA} and Lemma \ref{lemma:grads_SA_X} we compute the gradients with respect to the queries, values and $\Xm$, respectively. Then we use these results to prove Lemma \ref{lemma:gradients_queries} by computing the expectation of the Frobenius norms. 

\begin{tcolorbox}
\begin{lemma}[Gradients of Self Attention for parameter matrices]
\label{lemma:grads_SA}
The gradients of the self attention layer defined in Eq.~\eqref{eq:self_att} have the following form:
\begin{align*}
     & \frac{\partial \Sm}{\partial \wV} =    \soft\left(\frac{\Xm\wQ{\wK}^\top \Xm^\top}{\sqrt{d_k}}\right)\Xm\, \kro \Im_{d_v} \, \\
     & \frac{\partial \Sm}{\partial \wQ} = \left(\Im_n \kro {\wV}^\top\Xm^\top\right) \frac{\partial \Am}{\partial \Mm} \left(\frac{\Xm\kro\Xm\wK}{\sqrt{d_k}}\right),
\end{align*}
where the gradients of the softmax with respect to its inputs are as follows:
\begin{equation}
\label{eq:grad_soft_complete}
    \frac{\partial \Am}{\partial \Mm} = \blockdiag\Bigg( \dfrac{\partial \Am_{i}}{\partial \Mm_{i}^\top}\Bigg)
\end{equation}
and where $ \dfrac{\partial \Am_{i}}{\partial \Mm_{i}^\top}=\diag(\Am_{i}) - \Am_{i}\Am_{i}^\top$ with $\Am_{i}$ being the $i$-th row of $\Am$ in column vector format.\\
Finally, note that under the uniform-attention assumption, Eq.~\eqref{eq:grad_soft_complete} simplifies to:
\begin{equation}
\label{eq:grad_soft}
    \frac{\partial \Am}{\partial \Mm} = \frac{1}{n}\Im_n \kro \left(\Im_n - \frac{1}{n}\bm{1}_{n\times n} \right).
\end{equation}
\end{lemma}
\end{tcolorbox}

\begin{proof}
Let's start with the simple case of the values' weights $\wV$. Using the rule in Eq.~\eqref{eq:matrix-derivative}, it is immediate that:
\begin{equation*}
 \frac{\partial \Sm}{\partial \wV} =    \soft\left(\frac{\Xm\wQ{\wK}^\top \Xm^\top}{\sqrt{d_k}}\right)\Xm\, \kro \Im_{d_v} = \Am\Xm\, \kro \Im_{d_v} \,.
\end{equation*}
For the queries, a simple application of the chain rule and then again Eq.~\eqref{eq:matrix-derivative} gives:
\begin{align*}
    \frac{\partial \Sm}{\partial \wQ} &= \frac{\partial \Sm}{\partial \Am} \frac{\partial \Am}{\partial \wQ}
     = \frac{\partial \Sm}{\partial \Am} \frac{\partial \Am}{\partial \Mm} \frac{\partial \Mm}{\partial \wQ} \\
     &= \left(\Im_n \kro {\wV}^\top\Xm^\top\right) \frac{\partial \Am}{\partial \Mm} \left(\frac{\Xm\kro\Xm\wK}{\sqrt{d_k}}\right) \, ,
\end{align*}
which is the desired results. Finally, for the gradients of the softmax note that:
\begin{equation*}
    \frac{\partial{\Am_{pq}}}{\partial \Mm_{ij}} =
    \frac{\partial}{\partial \Mm_{ij}}\frac{\exp(\Mm_{pq})}{\sum_k \exp(\Mm_{pk})} =  \delta_{ip}\delta_{jq} \Am_{ij} - \delta_{ip} \Am_{iq}\Am_{ij} .
\end{equation*}
By writing the above expression in the matrix notation described above, we obtain the desired result. More specifically, the block diagonal structure is given from the term $\delta_{ip}$ which stems from the fact that the softmax is applied row-wise. 
\end{proof}
\begin{tcolorbox}

\begin{lemma}[Gradients of Self Attention with respect to the Embedding matrix]
\label{lemma:grads_SA_X}
The gradients of the self attention layer with respect to the embedding matrix $\Xm$ defined in Eq.~\eqref{eq:self_att} have the following form
\small{
\begin{align}
\label{eq:grad_inp}
    \frac{\partial\Sm}{\partial\Xm} 
    &= \frac{1}{\sqrt{d_k}}(\Im_n\kro \wVT\Xm^\top)\,\frac{\partial \Am}{\partial \Mm}  \,\left(\Im_n\kro\Xm\wK\wQT + \Km_{nn}(\Im_n\kro\Xm\wQ\wKT)\right) \,+ \,\Am\kro{\wV}^\top,
\end{align}}\normalsize
where the gradients of the softmax with respect to its inputs are denoted by $\frac{\partial \Am}{\partial \Mm}$ as before.
\end{lemma}
\end{tcolorbox}
\begin{proof}
Remember that we defined $\Sm = \soft(\extra{\frac{1}{\sqrt{d_k}}}\Xm\wQ\wKT\Xm^\top)\Xm\wV$. Alongside with our previous shorthands $\Am$, $\Mm$, let us define the remaining $\Xm\wV$ as a matrix $\Tm$, so that $\Sm=\Am\,\Tm$. Both $\Am$ and $\Tm$ are functions of $\Xm$. So the matrix differential can be written as:
\begin{align}
    \frac{\partial\Sm}{\partial\Xm} &=  \frac{\partial\Sm}{\partial\Am}\frac{\partial\Am}{\partial\Xm} + \frac{\partial\Sm}{\partial\Tm}\frac{\partial\Tm}{\partial\Xm}\\
    &=  \frac{\partial\Sm}{\partial\Am}\frac{\partial\Am}{\partial\Mm} \frac{\partial \Mm}{\partial\Xm} + \frac{\partial\Sm}{\partial\Tm}\frac{\partial\Tm}{\partial\Xm}\\
    &= (\Im_n\kro \wVT\Xm^\top)\,\frac{\partial \Am}{\partial \Mm}  \,\frac{\partial \Mm}{\partial \Xm} + (\Am\kro\Im_d)(\Im_n\kro\wVT)\\
    &= (\Im_n\kro \wVT\Xm^\top)\,\frac{\partial \Am}{\partial \Mm}  \,\frac{\partial \Mm}{\partial \Xm} + (\Am\kro\wVT)\label{eq:grad-SA-X}
\end{align}

Next, we use the matrix differential and then the identification theorem of matrix derivatives to compute the matrix gradient $\frac{\partial \Am}{\partial \Xm}$
\begin{align*}
    \mathrm{d} \Am &=  \extra{\frac{1}{\sqrt{d_k}}}\mathrm{d}(\Xm)\, \wQ\wKT\Xm^\top + \extra{\frac{1}{\sqrt{d_k}}}\Xm\wQ\wKT  \,\mathrm{d}(\Xm^\top).
\end{align*}

Vectorizing both sides:
\begin{align*}
    \mathrm{d} \vect_r(\Am) &=  \extra{\frac{1}{\sqrt{d_k}}}(\Im_n\kro\Xm\wK\wQT)\mathrm{d}(\vect_r(\Xm))\, + \,\extra{\frac{1}{\sqrt{d_k}}}(\Xm\wQ\wKT \kro\Im_n) \,\mathrm{d}(\vect_r(\Xm^\top)) \\
    &=  \extra{\frac{1}{\sqrt{d_k}}}(\Im_n\kro\Xm\wK\wQT)\mathrm{d}(\vect_r(\Xm))\, + \,\extra{\frac{1}{\sqrt{d_k}}}(\Xm\wQ\wKT \kro\Im_n) \Km_{dn}\,\mathrm{d}(\vect_r(\Xm)).
\end{align*}

Recall, for an arbitrary matrix $\Bm\in\mathbb{R}^{m\times n}$, the commutation matrix $\Km_{mn}$ transforms columnwise vectorization into rowwise vectorization. More precisely,
\begin{align*}
    \Km_{mn}\vect_c(\Bm) = \vect_c(\Bm^\top)
\end{align*}
and $\vect_c(\Bm) = \vect_r(\Bm^\top)$. Therefore, for rowwise vectorization, we have a similar result:
\begin{align*}
    \Km_{mn}\vect_r(\Bm^\top) &= \vect_r(\Bm)\\
    \vect_r(\Bm^\top) &= \Km_{nm}\vect_r(\Bm),
\end{align*}
where in the last line we used the fact the commutation is a permutation matrix, so $\Km_{mn}^{-1}=\Km_{mn}^\top=\Km_{nm}$. Thus, we get the required matrix derivative as follows:
$$\frac{\partial \Am}{\partial \Xm} = \extra{\frac{1}{\sqrt{d_k}}}\Im_n\kro\Xm\wK\wQT + \extra{\frac{1}{\sqrt{d_k}}}(\Xm\wQ\wKT\kro\Im_n)\Km_{dn}\,.$$
Next, we will use a property of commutation matrix to make things simpler (Theorem 7.9, \cite{magnus2019matrix}):
$$
\frac{\partial \Am}{\partial \Xm} = \extra{\frac{1}{\sqrt{d_k}}}\Im_n\kro\Xm\wK\wQT + \extra{\frac{1}{\sqrt{d_k}}}\Km_{nn}(\Im_n\kro\Xm\wQ\wKT).
$$
Plugging this into the above Eq.~\eqref{eq:grad-SA-X}, we get:
\begin{align*}
     \frac{\partial\Sm}{\partial\Xm} 
    &= \extra{\frac{1}{\sqrt{d_k}}}(\Im_n\kro \wVT\Xm^\top)\,\frac{\partial \Am}{\partial \Mm}  \,\left(\Im_n\kro\Xm\wK\wQT + \Km_{nn}(\Im_n\kro\Xm\wQ\wKT)\right) \,+ \,\Am\kro{\wV}^\top.
\end{align*}
As a sanity check, we can calculate if the shapes of the matrices are consistent. LHS should be a $nd\times nd$ matrix, while the constituent matrices of the first term on RHS: $\Im_n\kro {\wV}^\top\Xm^\top\in\mathbb{R}^{nd\times n^2}$, $\frac{\partial \Am}{\partial \Mm} \in\mathbb{R}^{n^2\times n^2}$, the additive term next to it is a $n^2\times nd$ matrix, and the second term on RHS is a Kronecker product of a $n\times n$ and a $d\times d$ matrix. 
\end{proof}

\begin{tcolorbox}
\gradients*
\end{tcolorbox}

\begin{proof}
Here, we suppress the index $\ell$.

\textbf{Gradient with respect to the values matrix.}

Recall that from Lemma \ref{lemma:grads_SA} we have that:
\begin{equation*}
    \frac{\partial \Sm}{\partial \wV} =    \soft\left(\frac{\Xm\wQ{\wK}^\top \Xm^\top}{\sqrt{d_k}}\right)\Xm\, \kro \Im_{d_v} \overset{\text{Ass. } \ref{ass:uniform_softmax}}{=} \frac{1}{n}\bm{1}_{n \times n} \Xm \kro \Im_{d_v}.
\end{equation*}
By direct computation:
\begin{align*}
    \norm{\frac{\partial \Sm}{\partial \wV}}_F^2 = \tr\left(\frac{\partial \Sm}{\partial \wV} \, \frac{\partial \Sm}{\partial \wV}^\top\right) \overset{\eqref{eq:prod_kro}}&{=} \frac{1}{n^2} \tr((\bm{1}_{n\times n} \Xm \Xm^\top \bm{1}_{n\times n}) \kro \Im_{d_v})\\
    \overset{\eqref{eq:trace_kro}}&{=}\frac{1}{n^2} \tr(\bm{1}_{n\times n} \Xm \Xm^\top \bm{1}_{n\times n})\tr(\Im_{d_v})\\
    &= \frac{d}{n^2}\tr( \Xm \Xm^\top \bm{1}_{n\times n}\bm{1}_{n\times n}) \\
    &= \frac{d_v}{n}\tr( \Xm \Xm^\top \bm{1}_{n\times n}) \\
    &= \frac{d_v}{n}\tr(\Xm\Xm^\top \bm{1}_n \bm{1}_n^\top) \\
    &= \frac{d_v}{n} \bm{1}_n^\top\Xm \Xm^\top \bm{1}_n \\
    &= \frac{d_v}{n} \|\Xm^\top\bm{1}_n\|^2=d_vn\norm{\bm{\bar{x}}}^2 .
\end{align*}

\textbf{Gradients with respect to the queries/keys matrix.}

First, recall the expression for the gradient of the softmax under the uniform-attention assumption (Eq.~\eqref{eq:grad_soft}):
\begin{equation*}
    \frac{\partial \Am}{\partial \Mm} = \frac{1}{ n}\Im_n \kro \left(\Im_n - \frac{1}{n}\bm{1}_{n\times n} \right).
\end{equation*}

Hence, we can rewrite the expression of Lemma \ref{lemma:grads_SA} for the gradients of the queries as:
\begin{align*}
\frac{\partial \Sm}{\partial \wQ} 
     &= \left(\Im_n \kro {\wV}^\top\Xm^\top\right) \frac{\partial \Am}{\partial \Mm} \left(\frac{\Xm\kro\Xm\wK}{\sqrt{d_k}}\right)\\ &= \frac{1}{\sqrt{d_k}n}\left(\Im_n \kro {\wV}^\top\Xm^\top\right) \left[\Im_n \otimes \left(\Im_n - \frac{1}{n}\bm{1}_{n\times n} \right)\right] \left(\Xm\kro\Xm\wK\right)\\
     &= \frac{1}{\sqrt{d_k}n}\Xm \kro \left[ {\wV}^\top\Xm^\top\left(\Im_n - \frac{1}{n}\bm{1}_{n\times n} \right)\Xm\wK\right],
\end{align*}
where in the last step we have used twice the property of the Kronecker product in Eq.~\eqref{eq:prod_kro} of Lemma~\ref{lemma:prop_kro}.

Hence,
\begin{align*}
\left\|\frac{\partial \Sm}{\partial \wQ}\right\|_F^2 = \tr \left(\frac{\partial \Sm}{\partial \wQ} \frac{\partial \Sm}{\partial \wQ} ^T\right) \overset{\eqref{eq:trace_kro}}{=} \frac{1}{d_k n^2} \|\Xm\|^2_F \cdot\left\|{\wV}^\top\Xm^\top\left(\Im_n - \frac{1}{n}\bm{1}_{n\times n} \right)\Xm\wK\right\|_F^2,
\end{align*}
where we have used the property on the trace of the Kronecker product (Lemma \ref{lemma:prop_kro}, Eq.~\eqref{eq:trace_kro}). 
Note that if we are conditioning on $\Xm$, then we only have to take the expectation of the last term with respect to the weights $\Wm^K$ and $\Wm^V$. Let us call $\Lm :=\Im_n - \frac{1}{n}\bm{1}_{n\times n}$ for notation simplicity.

Note: for a matrix $\Wm\in\mathbb{R}^{d\times d}$ whose entries $w_{ij}\sim\mathcal{N}(0, \sigma^2)$, then $\Exp \Wm\Wm^\top=d\sigma^2 \, \Im_d$. Thus, exchanging the order of trace and expectation, we can write: 
\begin{align*}
    \Exp\|\wVT\Xm^\top\Lm\Xm\wK\|^2_F &= \Exp\tr(\wVT\Xm^\top\Lm\Xm\wK \, \cdot\,\wKT\Xm^\top\Lm\Xm\wV)\\
    &=\tr(\Xm^\top\Lm\Xm\Exp[\wK\wKT]\Xm^\top\Lm\Xm\Exp[\wV\wVT])\\
    &=\sigma^2_v\sigma^2_k d_kd_v \tr(\Xm^\top\Lm\Xm\,\cdot\,\Xm^\top\Lm\Xm)\\
    &=\sigma^2_v\sigma^2_k d_kd_v\|\Xm^\top\Lm\Xm\|^2_F\\ &=\sigma^2_v\sigma^2_k d_kd_v\norm{\Xm^\top(\Im_n - \frac{1}{n}\bm{1}_{n}\bm{1}_{n}^\top)\Xm}^2_F \\
    &=\sigma^2_v\sigma^2_k d_kd_v\|\Xm^\top\Xm - n\bar{\bm{x}}\bar{\bm{x}}^\top\|^2_F,
\end{align*}
where, $\bar{\bm{x}}=\frac{1}{n}\Xm^\top\bm{1}_n\in\mathbb{R}^d$ is the mean embedding. Multiply this by $\frac{1}{d_kn^2} \|\Xm\|^2_F$ to get the final answer.

\textbf{Gradient with respect to the input.}

Plugging in the values of $\frac{\partial \Am}{\partial \Mm}$ and $\Am$ under the uniform-attention assumption into Eq.~\eqref{eq:grad_inp} gives rise to the following:
\begin{align*}
     \frac{\partial\Sm}{\partial\Xm} 
    &= \frac{1}{n\extra{\sqrt{d_k}}}\Im_n\kro \wVT\Xm^\top\left(\Im_n - \frac{1}{n}\bm{1}_{n}\bm{1}_{ n}^\top\right)\Xm\wK\wQT  \\& + \frac{1}{n\extra{\sqrt{d_k}}}\left[\Im_n\kro \wVT\Xm^\top \left(\Im_n - \frac{1}{n}\bm{1}_{n}\bm{1}_{ n}^\top\right)\right] \Km_{nn}(\Im_n\kro\Xm\wQ\wKT) \, \\&+
    \frac{1}{n}\bm{1}_{n\times n}\kro\wVT
\end{align*}
Let's refer to the matrices on the right-hand side as $\Am_1, \Am_2, \Am_3$ respectively. We compute the expected squared Frobenius norm of these as follows:

For $\Am_3$:
\begin{align*}
    \Exp[\norm{\Am_3}^2_F] &= \frac{1}{n^2}\Exp[\tr(n \bm{1}_{n\times n}\kro\wVT\wV)] \\
    \overset{\eqref{eq:trace_kro}}&{=}\frac{1}{n}\tr(\bm{1}_{n\times n})\tr(\Exp[\wV\wVT]) = d_v^2\sigma^2_v.
\end{align*}

Similarly, for $\Am_1$:
\begin{align}
    \Exp[\norm{\Am_1}^2_F] &= \frac{\sigma^2_q\sigma^2_k\sigma^2_v d_kd_v}{n^2} \, \tr(\Im_n)\tr(\Xm^\top\Lm\Xm \,\cdot\,\Xm^\top\Lm\Xm)\\
    &=\frac{\sigma^2_q\sigma^2_k\sigma^2_v d_kd_v}{n} \, \norm{\Xm^\top\Xm - n\bar{\bm{x}}\bar{\bm{x}}^\top}^2_F\\
    &=\sigma^2_q\sigma^2_k\sigma^2_v d_kd_v n \, \norm{\frac{1}{n}\Xm^\top\Xm - \bar{\bm{x}}\bar{\bm{x}}^\top}^2_F .
\end{align}
Finally, for $\Am_2$:
\begin{align}
    &\Exp[\norm{\Am_2}^2_F]\\
    &= \frac{1}{n^2 d_k}\Exp\left[\tr( \left[\Im_n\kro \wVT\Xm^\top \Lm \right] \Km_{nn}(\Im_n\kro\Xm\wQ\wKT\wK\wQT\Xm^\top) \Km_{nn}\left[\Im_n\kro \Lm\Xm\wV \right])\right]\\
    &= \frac{\sigma^2_q\sigma^2_k\sigma^2_v d_kd_v}{n^2}\,\tr((\Im_n\kro\Xm\Xm^\top)[\Im_n \kro \Lm\Xm\Xm^\top\Lm]) \\
    &= \frac{\sigma^2_q\sigma^2_k\sigma^2_v d_kd_v}{n}\,\tr(\Xm^\top\Lm\Xm \,\cdot\,\Xm^\top\Lm\Xm) =\Exp[\norm{\Am_1}^2_F],
\end{align}
where in the second line we have taken the expectation inside and used the fact that $\Km_{nn}$, being a commutation matrix, is orthogonal. Then, by simple properties of Kronecker product and cyclic property of trace, we have the result, which is the same as that for $\Am_1$. \\

Finally, by the triangle inequality
\begin{align}
    \Exp\left\|\frac{\partial\Sm}{\partial\Xm}\right\|^2 &\le 2 \Exp\|\Am_1+\Am_2\|^2 + 2 \Exp\|\Am_3\|^2 \\
    &\le 4 \Exp\|\Am_1\|^2 + 4 \Exp\|\Am_2\|^2+ 2\Exp\|\Am_3\|^2\\
    &= 8 \Exp\|\Am_1\|^2 + 2\Exp\|\Am_3\|^2\\
    &= \frac{8\sigma^2_q\sigma^2_k\sigma^2_v d_kd_v}{n}\norm{\Xm^\top\Xm - n\bar{\bm{x}}\bar{\bm{x}}^\top}^2_F + 2d_v^2\sigma_v^2.
\end{align}
This completes the proof.
\end{proof}

\subsubsection{Proof of Theorem \ref{thm:vanishing_gradients}}

\begin{tcolorbox}
\vanishinggradients*
\end{tcolorbox}

Before starting the proof, it is interesting to note that, even though the gradients of queries and keys vanish in the rank collapse regime~(i.e. $\norm{\Xm^\top\Xm - n\bar{\bm{x}}\bar{\bm{x}}^\top}=0$), the gradient with respect to the values and the input does not~(see Theorem~\ref{lemma:gradients_queries}). From this simple remark, we can conclude that, even in the rank collapse regime, information still propagates in the backward pass. In Section~\ref{sec:adaptive}~(main paper), we show that even if gradients effectively propagate, the phenomenon studied in this theorem still greatly affects training.

\begin{proof}
By using the chain rule and the fact that for two matrixes $\Am, \Bm$ we have that $\norm{\Am\Bm}_F^2 \leq \norm{\Am}_F^2\norm{\Bm}_F^2$, we can upper bound the gradient as:
\begin{align*}
    \norm{\frac{\partial \mathcal{L}}{\partial{\Wm^{Q,\ell}}}}_F^2 &\leq \prod_{i=\ell+1}^{L-1} \norm{\frac{\partial \Xm^{i+1}}{\partial \Xm^i}}_{F}^2 \norm{\frac{\partial \mathcal{L}}{\partial \Xm^L}}_F^2 \norm{\frac{\partial \Xm^{\ell+1}}{\partial \Wm^{Q,\ell}}}_F^2 \\
    &\leq \prod_{i=\ell+1}^{L-1} \norm{\frac{\partial \Xm^{i+1}}{\partial \Xm^i}}_{F}^2 \norm{\frac{\partial \mathcal{L}}{\partial \Xm^L}}_F^2 \norm{\frac{\partial \Xm^{\ell+1}}{\partial \Zm^{\ell}}}_F^2 \norm{\frac{\partial \Zm^\ell}{\partial \Wm^{Q, \ell}}}_F^2 \\
    &\leq \prod_{i=\ell+1}^{L-1} \norm{\frac{\partial \Xm^{i+1}}{\partial \Xm^i}}_{F}^2 \norm{\frac{\partial \mathcal{L}}{\partial \Xm^L}}_F^2 \norm{\frac{\partial \Xm^{\ell+1}}{\partial \Zm^{\ell}}}_F^2 \left(\norm{\frac{\partial \alpha_1 \Sm^{\ell}}{\partial \Wm^{Q, \ell}}}_F^2 + \underbrace{\norm{\frac{\partial \Xm^{\ell}}{\partial \Wm^{Q, \ell}}}_F^2}_{=0} \right) ,
\end{align*}
where we recall that $\Zm^{\ell} = \alpha_1 \Sm^{\ell}+ \Xm^{\ell}$ and in the last step we have used that $\Xm^\ell$ does not depend on $\Wm^{Q,\ell}$, hence the gradient vanishes. By taking expectation and using the tower property, we have that:
\begin{equation*}
    \Exp \norm{\frac{\partial \mathcal{L}}{\partial{\Wm^{Q,\ell}}}}_F^2 \leq \Exp\left[\underbrace{\Exp\left[\prod_{i=\ell+1}^{L-1} \norm{\frac{\partial \Xm^{i+1}}{\partial \Xm^i}}_{F}^2 \norm{\frac{\partial \mathcal{L}}{\partial \Xm^L}}_F^2 \norm{\frac{\partial \Xm^{\ell+1}}{\partial \Zm^{\ell}}}_F^2\right]}_{=:G(\Xm^\ell)} \norm{\frac{\partial \alpha_1 \Sm^{\ell}}{\partial \Wm^{Q, \ell}}}_F^2   \right],
\end{equation*}
where the expectations are taken with respect to $\Xm^\ell$ for the outer one and conditioning on $\Xm^\ell$ for inner one. Indeed, the first three terms only depend on the network values after $\Xm^\ell$. Now, a repeated application of the tower property in $G(\Xm^\ell)$, together with the results on the gradients of Lemma \ref{lemma:gradients_queries}, easily shows that $G(\Xm^\ell)$ stays bounded under our hypothesis. To see this one can also simply note that, since the softmax and its derivatives are almost surely bounded, the boundedness of $G(\Xm^\ell)$ is implied by an analogous statement for a vanilla linear MLP~(i.e removing the softmax). In this setting, the random variable inside the expectation in $G(\Xm^\ell)$ is a finite linear combination of Gaussian products --- which has bounded expectation.

All in all, we have that
\begin{equation*}
    \Exp \norm{\frac{\partial \mathcal{L}}{\partial{\Wm^{Q,\ell}}}}_F^2\le \Exp\left[ B_{\Xm^\ell}\norm{\frac{\partial \alpha_1 \Sm^{\ell}}{\partial \Wm^{Q, \ell}}}_F^2\right],
\end{equation*}
where $B_{\Xm^\ell}$ is an almost-surely-bounded function of $\Xm^{\ell}$. Hence, to show that $\Exp \norm{\frac{\partial \mathcal{L}}{\partial{\Wm^{Q,\ell}}}}_F^2=0$, we now just need to show that:
\begin{equation*}
    \Exp \norm{\frac{\partial \alpha_1 \Sm^{\ell}}{\partial \Wm^{Q, \ell}}}_F^2 = 0
\end{equation*}
under the rank-1 hypothesis for $\Xm^\ell$.
Let $\Xm_{1}^\ell, \dots \Xm_{n}^\ell \in \mathbb{R}^{d_v}$  be the representations for the $n$ tokens. Under the rank-1 assumption, each token can be written as a multiple of a single vector $\bm{x} \in \mathbb{R}^{d_v}$, and hence there exists $a_1, \dots, a_n \in \mathbb{R}$ such that $\Xm_1 = a_1 \bm{x}, \dots, \Xm_n = a_n \bm{x}$. From Lemma \ref{lemma:gradients_queries}, we know that:
\begin{equation*}
    \mathbb{E}\left\|\frac{\partial \Sm^{\ell}}{\partial \wQ} \right\|^2_F = \frac{\sigma^2_v\sigma^2_k d^2}{n^2}\cdot \Exp \left[ \|\Xm^{\ell}\|^2_F \cdot  \|(\Xm^{\ell})^\top\Xm^{\ell} - n\bar{\bm{x}}^{\ell}(\bar{\bm{x}}^{\ell})^\top\|^2_F\right] .
\end{equation*}
The mean token simplifies to $\bar{\bm{x}}^l = \frac{\bm{x}}{n}\sum_k a_k$ and hence $\left(\bar{\bm{x}}^{\ell}(\bar{\bm{x}}^{\ell})^\top\right)_{ij} = \frac{1}{n^2} (\sum_{k}a_k)^2 x_ix_j$. Similarly, $\left((\Xm^{\ell})^\top\Xm^{\ell}\right)_{ij} = \sum_k a_k^2 x_i x_j$. If furthermore all the coefficients $a_i$ are the same (which corresponds to the rank collapse assumption $\Xm^{\ell}=\bm{1}_{n}\bm{x}^T$ analyzed here), then it is easy to see that $\left((\Xm^{\ell})^\top\Xm^{\ell}\right)_{ij} - n \left(\bar{\bm{x}}^{\ell}(\bar{\bm{x}}^{\ell})^\top\right)_{ij} = 0 \; \forall i,j$ and hence $\|(\Xm^{\ell})^\top\Xm^{\ell} - n\bar{\bm{x}}^{\ell}(\bar{\bm{x}}^{\ell})^\top\|^2_F = 0$.
\end{proof}

\subsection{Gradient Analysis of Section \ref{sec:dep_angle}}
\label{sec:constan_cosine_analysis}

Throughout this section we assume that between every pair of tokens, the same dimension is a zero-mean Gaussian random variable with the same correlation, meaning that
\begin{equation*}
    \Exp\left[\Xm_{i,j}\Xm_{i',j'}\right] = \begin{cases}
    0 & j\ne j' \ \ \text{(independent dimensions)}\\
    \sigma^2_x & i=i', j=j'\\
    \rho\sigma^2_x & i\ne i', j=j'.
    \end{cases}
\end{equation*}

As we will deal with the computation of 4-th order moments of correlated Gaussian random variables, we will make use of Isserlis theorem \citep{isserlis1918formula}:
\begin{tcolorbox}
    \begin{theorem}[Isserlis]
    Let $X_1, \dots X_m$ be $m$ zero-mean Gaussian random variables. Then:
    \begin{equation}
        \Exp[X_1 \cdots X_m] = \begin{cases}
    \sum_{p\in P_m^2} \prod_{(i,j) \in p} \Exp[X_iX_j] & m \text{ even} \\
    0 & m \text{ odd}
    \end{cases} 
    \end{equation}
    
    where $P_m^2$ is the set of all the possible pairings of the indexes $1,\dots, m$. 
    \end{theorem}
\end{tcolorbox}
In particular, we will only need the 4-th order term, which reads:
\begin{equation*}
    \Exp[X_1X_2X_3X_4] = \Exp[X_1X_2]\Exp[X_3X_4] + \Exp[X_1X_3]\Exp[X_2X_4] + \Exp[X_1X_4]\Exp[X_2X_3].
\end{equation*}

Now we can prove Eq.~\eqref{eq:grad_V} , which we re-state here:
\begin{tcolorbox}
\begin{equation*}
    \Exp \norm{\frac{\partial \Sm}{\partial \wV}}_F^2 = \sigma_x^2 d^2 \left(1 + \rho (n-1)  \right).
\end{equation*}
\end{tcolorbox}
Also, from Eq.~\eqref{eq:jacobian_values} we have that 
\begin{equation*}
    \mathbb{E}\left\|\frac{\partial \Sm^{\ell}}{\partial \Wm^{V,\ell}}\right\|^2_F = dn \mathbb{E}\|\bar{\bm{x}}^{\ell}\|^2.
\end{equation*}

Now,
\begin{equation*}
    \mathbb{E}\|\bar{\bm{x}}^{\ell}\|^2 = \mathbb{E} \left( \sum_{i=1}^d (\bar{\bm{x}}^{\ell}_i)^2 \right).
\end{equation*}
Each $\bar{\bm{x}}^{\ell}_i = \frac{1}{n} \sum_{k=1}^n \Xm^\ell_{ki}$ is equally distributed with mean 
\begin{equation*}
    \mathbb{E} [\bar{\bm{x}}^{\ell}_i] = \mathbb{E} \left[\frac{1}{n} \sum_{k=1}^n \Xm^\ell_{ki}\right] = 0
\end{equation*}
and variance
\begin{equation*}
    \text{Var} [\bar{\bm{x}}^{\ell}_i] = \text{Var} \left[\frac{1}{n} \sum_{k=1}^n \Xm^\ell_{ki}\right] = \frac{1}{n^2} \left(n\sigma_x^2 + n(n - 1) \rho \sigma_x^2 \right) = \frac{1}{n} \sigma_x^2 (1 + \rho (n - 1)).
\end{equation*}
Finally we get
\begin{equation*}
    \mathbb{E}\left\|\frac{\partial \Sm^{\ell}}{\partial \Wm^{V,\ell}}\right\|^2_F = \sigma_x^2 d^2 (1 + \rho (n - 1)).
\end{equation*}

We know prove Eq.~\eqref{eq:grad_Q}, which reads:
\begin{tcolorbox}
\begin{equation*}
    \Exp \left\|\frac{\partial \Sm}{\partial \wQ}\right\|_F^2 = \sigma_x^6 \frac{(n-1)}{n} (1 - \rho)^2 d (n + d) .
\end{equation*}
\end{tcolorbox}

For the queries (and the keys respectively), recall from Eq.~\eqref{eq:jacobian_queries} that
\begin{equation*}
    \mathbb{E}\left\|\frac{\partial \Sm^{\ell}}{\partial \Wm^{Q,\ell}} \right\|^2_F = \frac{\sigma^2_v\sigma^2_k d^2}{dn^2}\cdot \Exp \left[ \|\Xm^{\ell}\|^2_F \cdot  \|(\Xm^{\ell})^\top\Xm^{\ell} - n\bar{\bm{x}}^{\ell}(\bar{\bm{x}}^{\ell})^\top\|^2_F\right].  
\end{equation*}
To proceed, we drop the superscript $\ell$ and we make the additional assumption that $\|\Xm\|^2_F$ is uncorrelated from the correlation magnitude $\|\Xm^\top\Lm\Xm\|^2_F = \|\Xm^\top\Xm - n\bar{\bm{x}}\bar{\bm{x}}^\top\|^2_F$.

Let us proceed with an expansion:
\begin{align*}
    \Exp\left[\|\Xm^\top\Lm\Xm\|^2_F\right] &= \sum_{i,j=1}^d\Exp\left[\left(\sum_{a,b=1}^n \Xm_{ai} \Lm_{ab} \Xm_{bj}\right)^2\right]\\
    &= \sum_{i,j=1}^d\sum_{a,b, a', b'=1}^n  \Lm_{ab}\Lm_{a'b'}\Exp\left[\Xm_{ai}\Xm_{a'i} \Xm_{bj}\Xm_{b'j}\right].\\
\end{align*}
Now, we have 2 cases: if $i\ne j$, which gives $d(d-1)$ equal terms, we need to compute
\begin{equation*}
    \text{A} := \sum_{a,b, a', b'=1}^n  \Lm_{ab}\Lm_{a'b'}\Exp\left[\Xm_{ai}\Xm_{a'i}\right]\cdot \Exp\left[\Xm_{bj}\Xm_{b'j}\right],
\end{equation*}
where $(i,j)$ is any tuple with $i\ne j$ and we used uncorrelation of different dimensions. Otherwise, we get $d$ each equal to
\begin{equation*}
    \text{B} := \sum_{a,b, a', b'=1}^n  \Lm_{ab}\Lm_{a'b'}\Exp\left[\Xm_{ai}\Xm_{a'i}\Xm_{bi}\Xm_{b'i}\right],
\end{equation*}
where $i$ is any index.

\paragraph{Term A.} Note that
\begin{equation*}
    \Exp\left[\Xm_{ai}\Xm_{a'i}\right]\cdot \Exp\left[\Xm_{bj}\Xm_{b'j}\right]= \begin{cases}
     \sigma_x^4 & a=a', b=b',\quad n^2\  \text{terms}\\
     \rho\sigma_x^4 & a=a', b\ne b',\quad  n^2(n-1) \ \text{terms}\\
     \rho\sigma_x^4 & a\ne a', b=b',\quad  n^2(n-1) \ \text{terms}\\
     \rho^2\sigma_x^4 & a\ne a', b\ne b',\quad  n^2(n-1)^2 \ \text{terms}
    \end{cases}.
\end{equation*}
So basically $A$ is the sum of 3 terms:
\begin{align*}
    &A_1 := \sigma_x^4\sum_{a,b} \Lm_{ab}^2 = (n-1)\sigma_x^4 \\
    &A_2 := 2\rho\sigma_x^4\sum_{a,b}\sum_{b'\ne b}  \Lm_{ab}\Lm_{ab'} = -2(n-1)\rho\sigma_x^4 \\
    &A_3 := \rho^2\sigma^4_x \sum_{a,b}\sum_{a'\ne a}\sum_{b'\ne b}\Lm_{ab}\Lm_{a'b'} = \rho^2\sigma^4_x (n-1),
\end{align*}
where we leveraged the following direct calculations:
\begin{align*}
    A_1 &= \sigma_x^4\sum_{a,b} \Lm_{ab}^2\\
    &= \sigma_x^4\left(n\left(\frac{n-1}{n}\right)^2 + (n-1)n\frac{1}{n^2}\right)\\
    &= \sigma_x^4\frac{(n-1)^2+(n-1)}{n}\\
    &= \sigma_x^4(n-1).
\end{align*}
Next, we compute
\begin{align*}
    A_2 &= 2\rho\sigma_x^4\sum_{a,b}\Lm_{ab}\sum_{b'\ne b}  \Lm_{ab'}\\
    &= 2\rho\sigma_x^4\left(\sum_{a}\Lm_{a,a}\sum_{b'\ne b}  \Lm_{ab'} + \sum_{a}\sum_{b\ne a}\Lm_{ab}\sum_{b'\ne b}  \Lm_{ab'}\right)\\
    &= 2\rho\sigma_x^4\left(\sum_{a}\frac{n-1}{n}\left[-(n-1)\frac{1}{n}\right] + \sum_{a}\sum_{b\ne a}\Lm_{ab}\left[\frac{n-1}{n} - \frac{n-2}{n}\right]\right)\\
    &=2\rho\sigma_x^4\left(-\frac{(n-1)^2}{n} - \frac{1}{n}(n-1)n\frac{1}{n}\right)\\
    &= -2\rho\sigma_x^4(n-1).
\end{align*}
Finally, similar computations also lead to the last term. To follow the calculations, we invite the reader to draw the matrix $\Lm$ and to hide the columns over which summations are not performed:
\begin{align*}
    A_3 &=\rho^2\sigma^4_x \sum_{a,b}\Lm_{ab}\sum_{a'\ne a}\sum_{b'\ne b}\Lm_{a'b'}\\
    &= \rho^2\sigma^4_x \left(\sum_{a}\Lm_{aa}\sum_{a'\ne a}\sum_{b'\ne a}\Lm_{a'b'}+\sum_{a}\sum_{b\ne a}\Lm_{ab}\sum_{a'\ne a}\sum_{b'\ne b}\Lm_{a'b'}\right)\\
    &= \rho^2\sigma^4_x \left(\sum_{a}\Lm_{aa} (1 - \frac{1}{n}+\sum_{a}\sum_{b\ne a}\Lm_{ab} (-1\frac{1}{n}) \right)  \\
    &= \rho^2\sigma^4_x \left(n (1 - \frac{1}{n}) (1 - \frac{1}{n}) + n (n - 1) (- \frac{1}{n}) (- \frac{1}{n}) \right) \\
    &=\rho^2\sigma^4_x (n-1).
\end{align*}

All in all, we get:
\begin{equation*}
    \text{A} = (n-1)(1-\rho)^2\sigma^4_x.
\end{equation*}

\paragraph{Term B.}
We make use of Isserlis theorem, stating that:
\begin{equation*}
    \Exp\left[\Xm_{ai}\Xm_{a'i}\Xm_{bi}\Xm_{b'i}\right] = \underbrace{\Exp \Xm_{ai} \Xm_{a'i}\Exp\Xm_{bi}\Xm_{b'i}}_{Q_1} + \underbrace{\Exp \Xm_{ai} \Xm_{bi}\Exp\Xm_{a'i}\Xm_{b'i}}_{Q_2} + \underbrace{\Exp \Xm_{ai} \Xm_{b'i}\Exp\Xm_{a'i}\Xm_{bi}}_{Q_3}.
\end{equation*}

By using our independence assumptions, we get:
\begin{equation*}
    Q_1 = \sigma_x^4(\delta_{aa'} + \rho \delta_{a\neq a'})(\delta_{bb'} + \rho \delta_{b\neq b'}) = \sigma_x^4(\delta_{aa'}\delta_{bb'} + \rho \delta_{aa'}\delta_{b\neq b'} + \rho \delta_{a\neq a'}\delta_{bb'} + \rho^2 \delta_{a\neq a'}\delta_{b\neq b'}).
\end{equation*}
Similarly for $Q_2$ and $Q_3$:
\begin{equation*}
    Q_2 =\sigma_x^4(\delta_{ab}\delta_{a'b'} + \rho \delta_{ab}\delta_{a'\neq b'} + \rho \delta_{a\neq b}\delta_{a'b'} + \rho^2 \delta_{a\neq b}\delta_{a'\neq b'})
\end{equation*}
and
\begin{equation*}
    Q_3 = \sigma_x^4(\delta_{ab'}\delta_{a'b} + \rho \delta_{ab'}\delta_{a'\neq b} + \rho \delta_{a\neq b'}\delta_{a'b} + \rho^2 \delta_{a\neq b'}\delta_{a'\neq b}).
\end{equation*}

Hence,
\begin{equation*}
    B = \sum_{a,b, a', b'=1}^n  \Lm_{ab}\Lm_{a'b'}\Exp\left[\Xm_{ai}\Xm_{a'i}\Xm_{bi}\Xm_{b'i}\right] = \sum_{a,b, a', b'=1}^n  \Lm_{ab}\Lm_{a'b'}(Q_1 + Q_2 + Q_3).
\end{equation*} 

Let's study it term by term. We will also use $\Lm_{ab} = (\delta_{ab} - \frac{1}{n})$, and so $\Lm_{ab}\Lm_{a'b'} = (\delta_{ab}\delta_{a'b'} - \frac{\delta_{ab}}{n} - \frac{\delta_{a'b'}}{n} + \frac{1}{n^2})$.

\textbf{First term}: we have that $\sigma_x^4 \sum_{aa'b'b'} \Lm_{ab}\Lm_{a'b'} Q_1 $ which is equal to (omitting the constant $\sigma_x^4$):
\begin{align*}
    &=\sum_{a,a',b,b'} (\delta_{ab}\delta_{a'b'} - \frac{\delta_{ab}}{n} - \frac{\delta_{a'b'}}{n} + \frac{1}{n^2}) (\delta_{aa'}\delta_{bb'} + \rho \delta_{aa'}\delta_{b\neq b'} + \rho \delta_{a\neq a'}\delta_{bb'} + \rho^2 \delta_{a\neq a'}\delta_{b\neq b'}) \\
    &= \rho^2\left(n(n-1) - 2(n-1)(n-1) + (n-1)^2\right) + \rho (- 4(n-1) + 2(n-1) ) + n - 2 + 1 \\
    &= \rho^2(n-1)\left(n - 2(n-1) + (n-1)\right) - 2\rho(n-1) + (n - 1) \\
    &= \rho^2(n-1) - 2\rho(n-1) + (n-1).
\end{align*}

\textbf{Second term}: we have that $\sigma_x^4 \sum_{aa'b'b'} \Lm_{ab}\Lm_{a'b'} Q_2 $ which is equal to (omitting the constant $\sigma_x^4$):
\begin{align*}
    &=\sum_{a,a',b,b'}(\delta_{ab}\delta_{a'b'} - \frac{\delta_{ab}}{n} - \frac{\delta_{a'b'}}{n} + \frac{1}{n^2})(\delta_{ab}\delta_{a'b'} + \rho \delta_{ab}\delta_{a'\neq b'} + \rho \delta_{a\neq b}\delta_{a'b'} + \rho^2 \delta_{a\neq b}\delta_{a'\neq b'}) \\
    &= \rho^2(n-1)^2 + \rho(-2n(n-1) + 2(n-1)) + n^2 - 2n + 1 \\
    &= \rho^2 (n-1)^2 - 2\rho(n-1)^2 + (n-1)^2.
\end{align*}

\textbf{Third term:}
we have that $\sigma_x^4 \sum_{aa'b'b'} \Lm_{ab}\Lm_{a'b'} Q_3 $ which is equal to (omitting the constant $\sigma_x^4$):
\begin{align*}
    &=\sum_{a,a',b,b'}(\delta_{ab}\delta_{a'b'} - \frac{\delta_{ab}}{n} - \frac{\delta_{a'b'}}{n} + \frac{1}{n^2})(\delta_{ab'}\delta_{a'b} + \rho \delta_{ab'}\delta_{a'\neq b} + \rho \delta_{a\neq b'}\delta_{a'b} + \rho^2 \delta_{a\neq b'}\delta_{a'\neq b}) \\
    &= \rho^2\left( n(n-1) - 2(n-1)(n-1) + (n-1)^2\right) + \rho(-4(n-1) + 2(n-1)) + n - 2 + 1 \\
    &= \rho^2(n-1) - 2\rho(n-1) + (n - 1).
\end{align*}

Summing all the three terms, we get:
\begin{align*}
    B = \sigma_x^4(n-1)\left[ \rho^2 (n+1) - 2(n+1)\rho + (n+1) \right] = \sigma_x^4(n-1)(n + 1)(1 - \rho)^2.
\end{align*}

\textbf{Plugging in the values of A and B} we get:
\begin{equation*}
    \Exp[\|\Xm^\top \Lm\Xm\|^2_F] = d\cdot B + d(d-1)\cdot A = \sigma^4_x (1 - \rho)^2 d (n - 1) (n + d),
\end{equation*}
and finally assuming Xavier initialization
\begin{align*}
    \mathbb{E}\left\|\frac{\partial \Sm^{\ell}}{\partial \Wm^{Q,\ell}} \right\|^2_F &= \frac{\sigma^2_v\sigma^2_k d^2}{dn^2}\cdot \Exp \left[ \|\Xm^{\ell}\|^2_F \cdot  \|(\Xm^{\ell})^\top\Xm^{\ell} - n\bar{\bm{x}}^{\ell}(\bar{\bm{x}}^{\ell})^\top\|^2_F\right] \\
    &= \sigma_x^6 \frac{n - 1}{n} (1 - \rho)^2 d (n + d).
\end{align*}

\subsection{Forward Pass: Proofs of Lemma \ref{lemma:propagation_of_inner_producets} and \ref{thm:forward_pass} }
\label{app:forward_pass}
First, we characterize the evolution of the correlations between tokens $\Xm_{k}, \Xm_{k'}$ with depth, under the assumptions of Theorem \ref{thm:forward_pass}, namely uniform-attention assumption, and the adoption of a linear activation.


\begin{tcolorbox}
\begin{lemma}[Expectation of Linear Layers]
\label{lemma:exp_linear}
Let $\Dm = \Xm \Wm$, where $\Wm\in\mathbb{R}^{d\times d}$ is a random matrix with i.i.d random entries with variance $\sigma^2 = \frac{1}{d}$ and $\Xm\in\mathbb{R}^{n\times d}$ is a fixed matrix:
\begin{equation*}
    \Exp[\Dm_{kj}\Dm_{k'j}] = \frac{1}{d}\langle \Xm_k, \Xm_{k'} \rangle
\end{equation*}
\end{lemma}
\end{tcolorbox}
Note that by summing over the indexes, Lemma \ref{lemma:exp_linear} implies:
\begin{align*}
    &\Exp \norm{\Dm}_F^2 = \Exp \norm{\Xm}_{F}^2 \\
    &\Exp C(\Dm) = \Exp C(\Xm).
\end{align*}

\begin{proof}
\begin{equation*}
     \Exp [\Dm_{kj} \Dm_{k'j}] = \sum_{zz'} \Xm_{kz} \Xm_{k'z'} \Exp[\Wm_{zj}\Wm_{z'j}] = \sigma^2 \sum_{z}\Xm_{kz}\Xm_{k'z} = \frac{1}{d}\langle \Xm_k, \Xm_{k'} \rangle.
\end{equation*}
   
\end{proof}

\begin{tcolorbox}
\begin{lemma}[Expectation of skip connection]
\label{lemma:exp_skip}
    Let $\Am, \Bm \in \mathbb{R}^{p \times q}$. Let $\Dm := \alpha\Am + \Bm$ with $\Exp[\Am | \Bm] = \bm{0}$ and $\alpha \in \mathbb{R}$. Then:
    \begin{equation}
        \Exp\left[\Dm_{ij}\Dm_{i'j}\right] = \alpha^2 \mathbb{E}[\Am_{ij}\Am_{i'j}] + \mathbb{E}[\Bm_{ij}\Bm_{ij'}] 
    \end{equation}
    holds for all $i,i' \in [p], j \in [q]$.
\end{lemma}
\end{tcolorbox}
Note that by summing over the indexes, Lemma \ref{lemma:exp_skip} implies:
\begin{align*}
    &\Exp \norm{\Dm}_F^2 = \alpha^2\Exp \norm{\Am}_{F}^2 + \Exp \norm{\Bm}_F^2 \\
    &\Exp C(\Dm) = \alpha^2 \Exp C(\Am) + \Exp C(\Bm).
\end{align*}

\begin{proof}
\begin{align*}
    \mathbb{E}[\Dm_{ij}\Dm_{i'j}] &= \mathbb{E}\left[(\alpha \Am_{ij} + \Bm_{ij})(\alpha \Am_{i'j} + \Bm_{i'j})\right]\\
    &= \mathbb{E}\left[\alpha^2 \Am_{ij}\Am_{i'j} + \alpha \Am_{ij}\Bm_{i'j} + \alpha \Am_{i'j}\Bm_{ij} + \Bm_{ij}\Bm_{i'j} \right] \\
    &= \alpha^2\mathbb{E}\left[ \Am_{kj}\Am_{i'j}\right] + \mathbb{E}\left[ \Bm_{ij}\Bm_{i',j} \right],
\end{align*}
where using iterated expectations $\alpha \Exp[\Am_{i'j}\Bm_{ij}] = \alpha \Exp[\Exp [\Am_{i'j} | \Bm] \Bm_{ij}]] = 0$ and identically $\alpha \Exp [\Am_{ij}\Bm_{i'j}] = 0$.
\end{proof}

\begin{tcolorbox}
\begin{lemma}[Expectation of Attention Layers]
\label{lemma:exp_softmax}
Under the uniform-attention assumption:
\begin{equation*}
    \Exp[\Sm_{kj}\Sm_{k'j}] = \frac{1}{d_vn^2}\Exp C(\Xm).
\end{equation*}
\end{lemma}
\end{tcolorbox}
In this case, by summing over the indexes we have that:
\begin{align*}
    &\Exp\norm{\Sm}_F^2 = \frac{\Exp C(\Xm)}{n} \\
    & \Exp C(\Sm) = \Exp C(\Xm).
\end{align*}
\begin{proof}
Note that under the uniform-attention assumption:
\begin{equation*}
    \Sm_{kj} = \frac{1}{n}\left(\bm{1}_{n\times n}\Xm \Wm^V\right)_{kj} = \frac{1}{n}\sum_{zi}\Xm_{zi}\Wm^V_{ij}.
\end{equation*}
Hence, using the fact that the weights are i.i.d with variance $\sigma_v^2=\frac{1}{d_v}$:
\begin{align*}
    \Exp[\Sm_{kj}\Sm_{k'j}] = \frac{\sigma_{v}^2}{n^2}\sum_{z,z'}\sum_i\Exp[\Xm_{zi} \Xm_{z'i}] = \frac{1}{d_vn^2}\sum_{k,k'}\langle\Xm_z, \Xm_{z'}\rangle = \frac{1}{d_vn^2}\Exp C(\Xm).
\end{align*}
\end{proof}
\begin{tcolorbox}
\propinnprod*
\end{tcolorbox}

\begin{proof}
First, note that for the residual blocks we have that $\Exp [Y^{\ell}_{kj} | Z^{\ell}_{k'j}] = 0$ due to the independence assumption on the feedforward weights, and similarly $\Exp [S^{\ell}_{kj} | X^{\ell}_{k'j}] = 0$. Hence, we can use Lemma \ref{lemma:exp_skip} in both the skip connections of the Transformer architecture.
Therefore, using Lemma \ref{lemma:exp_skip} (skip), Lemma \ref{lemma:exp_linear} (linear) and Lemma \ref{lemma:exp_softmax} (attention):
\begin{align*}
        &\mathbb{E}[C(\Xm^{\ell+1})] \\
        \overset{\text{skip}}&{=} \alpha_2^2\mathbb{E}C(\Ym^\ell) + \mathbb{E}C(\Zm^{\ell}) \\ 
        \overset{\text{linear}}&{=} \alpha_2^2\mathbb{E}C(\Zm^{\ell}) + \mathbb{E}C(\Zm^{\ell}) \\ 
        &= (\alpha_2^2 + 1)\mathbb{E}C(\Zm^{\ell}) \\
        \overset{\text{skip}}&{=} (\alpha_2^2 + 1)\left(\alpha_1^2\mathbb{E}C(\Sm^{\ell}) + \mathbb{E}C(\Xm^{\ell})\right) \\ 
        \overset{\text{attention}}&{=}  (\alpha_2^2 + 1)(\alpha_1^2 + 1) \mathbb{E}[C(\Xm^{\ell})] \\ 
        \overset{\text{unroll recurs.}}&{=} (\alpha_2^2 + 1)^{\ell+1}(\alpha_1^2 + 1)^{\ell+1}C(\Xm) ,
    \end{align*}
    where in the last step we have unrolled the recursion until the input layer.
    
    For the limit as $L\to \infty$, simply note that:
    $$\lim_{L\to \infty}\left(\frac{\tilde{\alpha}_i}{L}+1\right)^{L} = \text{e}^{\tilde{\alpha}_i} ,$$ 
    with $i \in \{1, 2\}$.
\end{proof}

Now we are ready to re-state and prove Lemma \ref{thm:forward_pass}.
\begin{tcolorbox}
\forwardpass*
\end{tcolorbox}

\begin{proof}
The proof is in the same spirit as Lemma \ref{lemma:propagation_of_inner_producets} but slightly more involved. Again, using Lemma \ref{lemma:exp_skip} in both the skip connections of the Transfomer architecture.
Therefore, using Lemma \ref{lemma:exp_skip} (skip), Lemma \ref{lemma:exp_linear} (linear) and Lemma \ref{lemma:exp_softmax} (attention):
\begin{align*}
    \mathbb{E}[||\Xm^{\ell+1}||_F^2] \overset{\text{skip}}&{=} \alpha_2^2 \mathbb{E}||\Ym^{\ell}||_F^2 + \mathbb{E}||\Zm^{\ell}||_F^2 \\
    \overset{\text{linear}}&{=} (\alpha_2^2+1)\mathbb{E}||\Zm^{\ell}||_F^2 \\
    \overset{\text{skip}}&{=} (\alpha_2^2+1)\left(\alpha_1^2 \mathbb{E}[||\Sm^{\ell}||_F^2] + \mathbb{E}[||\Xm^{\ell}||_F^2] \right) \\
    \overset{\text{softmax}}&{=} (\alpha_2^2+1)\left(\frac{\alpha_1^2}{n} \mathbb{E}[C(\Xm^{\ell})] + \mathbb{E}[||\Xm^{\ell}||_F^2] \right) \\
    &= (\alpha_2^2+1)\frac{\alpha_1^2}{n} \mathbb{E}[C(\Xm^{\ell})] + (\alpha_2^2+1) \mathbb{E}[||\Xm^{\ell}||_F^2] \\
    \overset{\text{unroll }C(\Xm^\ell)}&{=} (\alpha_2^2+1)^{\ell+1}\alpha_1^2(\alpha_1^2+1)^{\ell}\frac{C(\Xm) }{n} + (\alpha_2^2+1) \mathbb{E}[||\Xm^{\ell}||_F^2] \\
    \overset{\text{unroll }\norm{\Xm^\ell}_F^2}&{=} (\alpha_2^2+1)^{\ell+1}\alpha_1^2\frac{C(\Xm) }{n} \sum_{k=0}^{\ell}(\alpha_1^2+1)^k + (\alpha_2^2+1)^{\ell+1} ||\Xm||_F^2,
\end{align*}

where in the second to last step we have used Lemma \ref{lemma:propagation_of_inner_producets} and in the last step we have unrolled the recursion for $\norm{\Xm^\ell}_F^2$ until the input layer. 

For the second part, we now show that for a network of $L$ layers, the choice $\alpha_1^2 =  \frac{\tilde{\alpha}_1}{L}$ and $\alpha_2^2 = \frac{\tilde{\alpha}_2}{L}$ stabilizes the norm of the activations in the forward pass. Using the product law for the limits, we can study the converges of $(\alpha_2^2+1)^{\ell}$ and $\alpha_1\sum_{k=0}^{\ell}(\alpha_1^2+1)^k$ separately. 

Let $i \in \{1, 2\}$. For the latter term we have that: 
\begin{align*}
    \lim_{L\to \infty} \frac{\tilde{\alpha}_i}{L}\sum_{l=0}^{L-1} \left(1 + \frac{\tilde{\alpha}_i}{L} \right)^{\ell} &= \lim_{L\to \infty} \frac{\tilde{\alpha}_i}{L}\frac{1-\left(1 + \frac{\tilde{\alpha}_i}{L} \right)^{\ell}}{1-1-\frac{\tilde{\alpha}_i}{L}} \\
    &= \lim_{L\to \infty} -1+\left(1 + \frac{\tilde{\alpha}_i}{L} \right)^{\ell}\\
    &= \text{e}^{\tilde{\alpha}_i} - 1  ,
\end{align*}
while for the former term we have that $\lim_{L\to \infty}(\frac{\tilde{\alpha}_i}{L}+1)^{\ell} = \text{e}^{\tilde{\alpha}_i}$. Hence, the norm of the representations converges to:
\begin{equation*}
    \lim_{L\to \infty}\mathbb{E}[||\Xm^{\ell}||_F^2] = \text{e}^{\tilde{\alpha}_2}(\text{e}^{\tilde{\alpha}_1} - 1)\frac{C(\Xm) }{n} + \text{e}^{\tilde{\alpha}_2} ||\Xm||_F^2.
\end{equation*}

The final results as stated in the theorem hold because of the following:

\textbf{Remark}: note that $C(\Xm)  = \sum_{k,k'}\sum_{j} \Xm_{kj}\Xm_{kj'} = \sum_{j} (\sum_{k,k'} \Xm_{kj}\Xm_{k'j}) = \sum_{j} (\sum_{k} \Xm_{kj})^2 = n^2 \norm{\bar{\bm{x}}}^2$.
\end{proof}

\subsection{Proof of Theorem \ref{thm:exp_cosine}: Correlations are Preserved under Residual Scaling}
\label{app:res_scaling_proofs}

\begin{tcolorbox}
\expectedcosine*
\end{tcolorbox}
\begin{proof}
Due to the rotational symmetries of the Gaussian random matrices, if the input  tokens have the same norm, then the expected norm at layer $\ell \in [L]$ is also the same across the token's representations. Hence, we can write $\Exp\norm{\Xm^\ell}_F^2 = n \Exp\norm{\bm{x}^\ell}^2$, where $\norm{\bm{x}^\ell}^2$ is the norm of every token at layer $\ell$. Furthermore, by definition of our correlation coefficient $\rho^l_{kk'}$, we have that $\Exp\langle\Xm^\ell_k, \Xm^\ell_{k'}\rangle = \rho^\ell_{kk'} \Exp\norm{\bm{x}^\ell}^2$. By summing over the indexes $k,k'$, we can expand the relation as:
\begin{equation*}
    \underbrace{\sum_{k,k'}\Exp\langle\Xm^\ell_k, \Xm^\ell_{k'}\rangle}_{\Exp C(\Xm)} = \sum_{k,k'} \rho^\ell_{kk'} \Exp\norm{\bm{x}^\ell}^2 = (n + \sum_{k\neq k'}\rho^\ell_{k,k'})\Exp\norm{\bm{x}^\ell}^2 = \underbrace{n\Exp\norm{\bm{x}^\ell}^2}_{\Exp \norm{\Xm^\ell}_F^2}(1 + (n-1) \rho^\ell).
\end{equation*}
By solving for $\rho^\ell$, we have that:
\begin{equation*}
    \rho^\ell = \frac{\Exp C(\Xm^\ell)}{(n-1)\Exp \norm{\Xm^\ell}^2 } - \frac{1}{n-1} .
\end{equation*}
Now we plug in the expressions for $\Exp C(\Xm^\ell)$ and $\Exp \norm{\Xm^\ell}^2 $ with the aid of Lemma \ref{lemma:propagation_of_inner_producets} and Lemma \ref{thm:forward_pass}, respectively. Finally, by taking the limits with respect to $L$, we get the desired result.
\end{proof}

\subsection{Motivation for Assumption~\ref{ass:uniform_softmax}}
\label{app:assumption_unif_soft}
We motivate here the following assumption, stated in the main paper. This assumption is crucial to compute expectations involving the softmax function.

\uniformsoftmax*

\paragraph{Theoretical analysis.}
We first show that this assumption holds when taking $d_k$ to infinity, keeping $d_{v}$ fixed.
\begin{tcolorbox}
\begin{lemma}
\label{app:convergence_A}
Consider initializing each entry of $\wQ\in\mathbb{R}^{d_{v}\times d_k}$ and $\wK\in\mathbb{R}^{d_{v}\times d_k}$ independently with variance $\sigma^2_k = 2/(d_{v}+d_k)$ --- i.e. Glorot initialization~\citep{glorot2010understanding}. Let $\Mm = \frac{1}{\sqrt{d_k}}\Xm^\ell\Wm^{Q,\ell}{\Wm^{K,\ell}}^\top{\Xm^\ell}^\top$; for any $(i,j)\in[n]\times[n]$ we have
\begin{equation}
    \Exp[\Mm_{i,j} \ | \ \Xm] = 0,\qquad \Exp[\Mm_{i,j}^2 \ | \ \Xm] = \sigma_k^4 \cdot \|\Xm_{i,:}\|^2 \cdot \|\Xm_{j,:}\|^2.
\end{equation}
While keeping $d_v<\infty$ fixed, taking $d_k$ to infinity yields
\begin{equation}
    \Exp[\Mm_{i,j}^2 \ | \ \Xm] = \mathcal{O}\left(\frac{1}{d_k^2}\right).
\end{equation}
In other words, $\Mm$ converges to $\bm{0}_{n\times n}$ in $L^2$ as $d_k\to\infty$. 
\end{lemma}
\end{tcolorbox}
\begin{proof}
First, note that
\begin{align*}
    \Mm_{i,j} = \frac{1}{\sqrt{d_k}}\sum_{a,c=1}^{d_{v}}\sum_{b=1}^{d_k} \Xm_{i,a}\wQ_{a,b}\wK_{c,b} \Xm_{j,c}.
\end{align*}
Since $\wQ$ is independent from $\wK$ at initialization, $\Exp[\Mm_{i,j} \ | \ \Xm] = 0$. Next, we compute
\begin{align*}
    \Exp[\Mm_{i,j}^2] &= \frac{1}{d_k}\sum_{a,c,a',c'=1}^{d_{v}}\sum_{b,b'=1}^{d_{k}} \Xm_{i,a} \Xm_{i,a'} \Xm_{j,c} \Xm_{j,c'} \Exp\left[\wQ_{a,b}\wQ_{a',b'}\wK_{c,b} \wK_{c',b'}\right] \\
    &=\frac{1}{d_k}\sum_{a,c,a',c'=1}^{d_{v}}\sum_{b,b'=1}^{d_{k}} \Xm_{i,a} \Xm_{i,a'} \Xm_{j,c} \Xm_{j,c'} \Exp\left[\wQ_{a,b}\wQ_{a',b'}\right]\Exp\left[\wK_{c,b} \wK_{c',b'}\right] \\
    &= \frac{\sigma_k^4}{d_k}\sum_{a,c=1}^{d_{v}}\sum_{b=1}^{d_{k}}  \Xm_{i,a}^2 \Xm_{j,c}^2\\
    &= \sigma_k^4 \|\Xm_{i,:}\|^2 \|\Xm_{j,:}\|^2.
\end{align*}
This concludes the proof.
\end{proof}

The following classical result implies almost sure convergence of the softmax matrix as $d_k\to\infty$.

\begin{lemma}[Borel-Cantelli]
Let $(X_i)$ be a sequence of random variables. If for any $\epsilon>0$
\begin{equation*}
    \sum_{i=0}^\infty \mathbb{P}[|X_i-X|>\epsilon]<\infty,
\end{equation*}
then $X_i$ converges to $X$ almost surely\footnote{That is, $\lim_{i\to\infty} X_i(\omega) = X(\omega)$ for almost every $\omega\in \Omega$~(i.e. with probability one).}.
\end{lemma}

\begin{tcolorbox}
\begin{theorem}[Almost-sure convergence]
\label{thm:soft_assumption_proof}
Consider initializing each entry of $\wQ\in\mathbb{R}^{d_v\times d_k}$ and $\wK\in\mathbb{R}^{d_v\times d_k}$ independently with variance $\sigma^2_k = 2/(d_v+d_k)$ --- i.e. Glorot initialization~\citep{glorot2010understanding}. Let $d_v<\infty$ be fixed, as $d_k\to\infty$ we have that, for any $\Xm$,
\begin{equation*}
    \Am := \soft\left(\frac{1}{\sqrt{d_k}}\Xm\Wm^{Q}{\Wm^{K}}^\top{\Xm}^\top\right)\stackrel{a.s.}{\to} \frac{1}{n}\bm{1}_{n\times n}
\end{equation*}
and
\begin{equation*}
    \frac{\partial\Am}{\partial\Mm} \stackrel{a.s.}{\to} \frac{1}{n}\Im_n \otimes \left(\Im_n - \frac{1}{n}\bm{1}_{n\times n} \right).
\end{equation*}
\end{theorem}
\end{tcolorbox}
\begin{proof}
Thanks to Lemma~\ref{app:convergence_A} and Markov Inequality, we have fast convergence in probability: for any fixed $\Xm$,
\begin{equation*}
    \mathbb{P}[|\Mm_{i,j}|>\epsilon]\le\frac{\Exp[\Mm^2_{i,j}]}{\epsilon^2} \le \frac{C_\epsilon}{d_k^2}.
\end{equation*}
Borel Cantelli then directly yields almost sure convergence of $\Mm$ to $\bm{0}_{n\times n}$ as $d_k\to\infty$. Next, note that both $\Am$ and $\frac{\partial\Am}{\partial\Mm}$ are continuous functions of $\Am$, hence we can apply standard continuity event-per-event. For almost every $\omega\in\Omega$,
\begin{equation*}
    \lim_{d_k\to\infty} \Am(\Am(\omega)) = \Am\left(\lim_{d_k\to\infty}\Am(\omega)\right) = \Am( \bm{0}_{n\times n}) = \frac{1}{n}\bm{1}_{n\times n}.
\end{equation*}
Hence $\Am\to \frac{1}{n}\bm{1}_{n\times n}$ almost surely. This can also be seen as a simple application of the continuous mapping theorem. The same reasoning yields almost sure convergence of
\begin{equation*}
    \frac{\partial \Am}{\partial \Mm} = \blockdiag\Bigg(\diag(\Am_{i:}) - \Am_{i:}\Am_{i:}^\top\Bigg),
\end{equation*}
to the corresponding limiting quantity.
\end{proof}

\vspace{10px}
\paragraph{Empirical analysis.}
\begin{wrapfigure}{r}{0.45\textwidth}
\vspace{-10px}
\includegraphics[scale = 0.4]{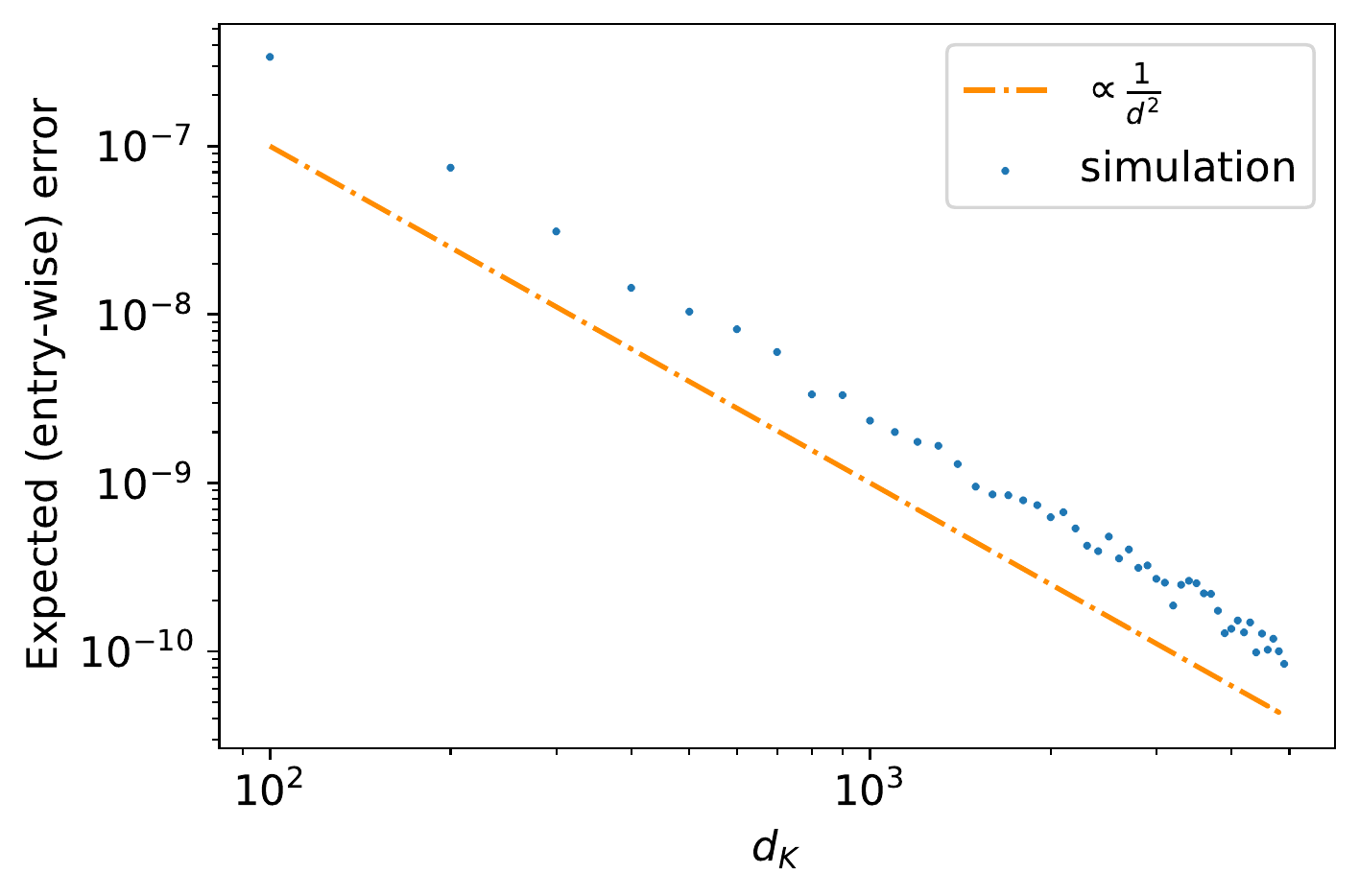}
\caption{\small{Evolution of $\frac{1}{n^2}\vert\vert\Am^{\ell}-\frac{1}{n}{\bf{1}}_{n\times n}\vert\vert_{F}^{2}$ as a function of $d_k$ for $d_v$ fixed at 100.}}\label{fig:softmaxuniform}
\vspace{-10px}
\end{wrapfigure} 
We empirically assess the validity of
Assumption~\ref{ass:uniform_softmax} and of its theoretical justification by performing the following experiments: for a range of increasing values of $d_k$, we compute $\Am$ and we calculate $\frac{1}{n^2}\vert\vert\Am^{\ell}-\frac{1}{n} {\bf{1}}_{n\times n}\vert\vert_{F}^{2}$, i.e. its average (entry-wise) distance from a uniform matrix with entries all equal to $1/n$. 
For each value of $d_k$, we repeat this calculation 200 times, each time with different random weight matrices. Fig.~\ref{fig:softmaxuniform} displays how the $\vert\vert\Am-\frac{1}{n}{\bf{1}}_{n\times n}\vert\vert_{F}^{2}$ averaged over 200 runs, tends to zero with a trend inversely proportional to $d_k^2$, as predicted by our theoretical analysis.

\vspace{20px}
\section{Additional Results}
\label{app:more_experiments}

\subsection{On the Roles of the $1/\sqrt{L}$-Scaling of the Residuals and Layer Normalization}
\begin{figure}
    \centering
    \includegraphics[scale=0.28]{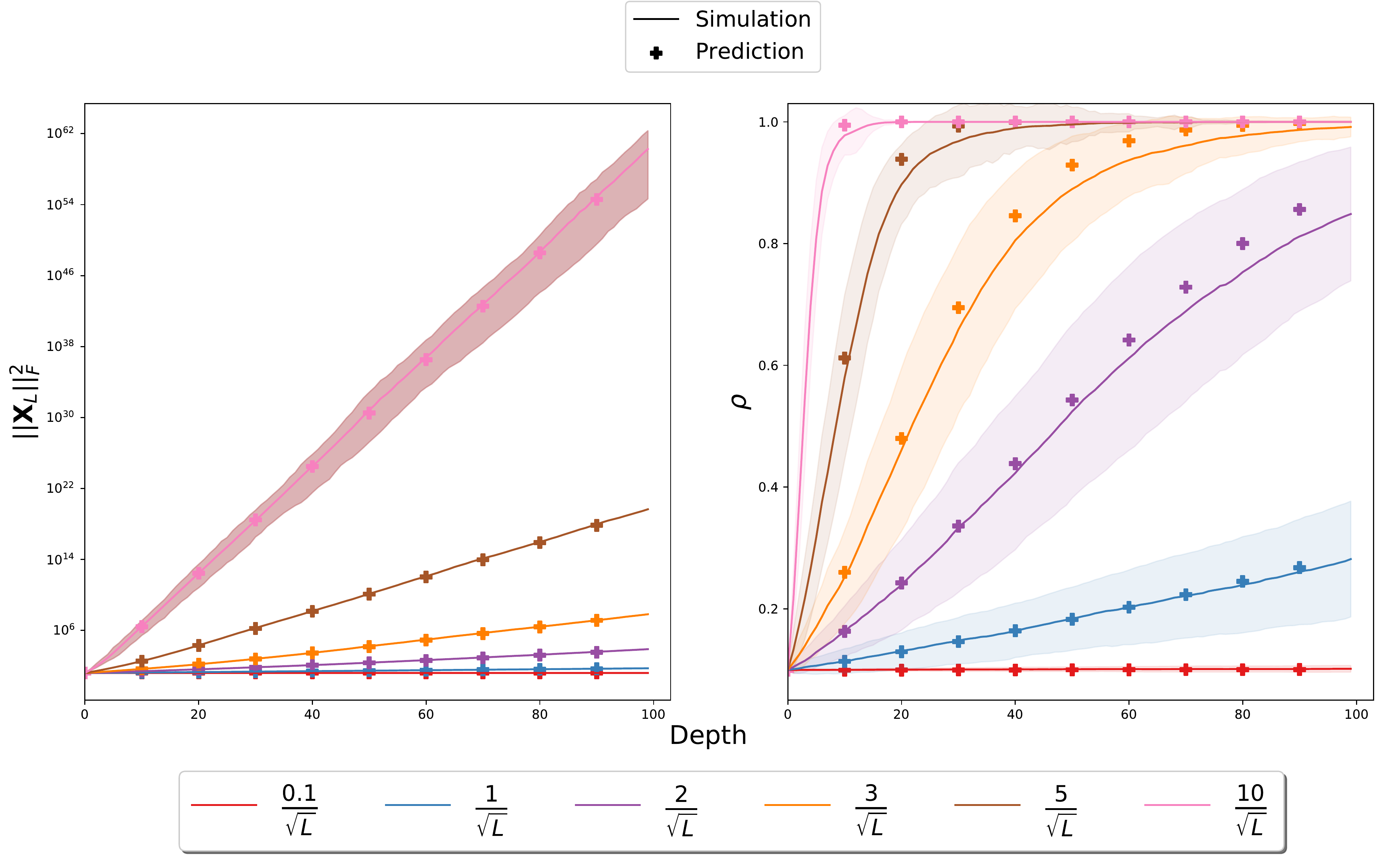}
    \caption{(Left) Propagation of the Frobenius norm of the input sequence; (Right) Propagation of the average token correlation.}
    \label{fig:norm_corr_prop_1}
\end{figure}

\begin{figure}[t]
    \centering
    \includegraphics[width=1\linewidth]{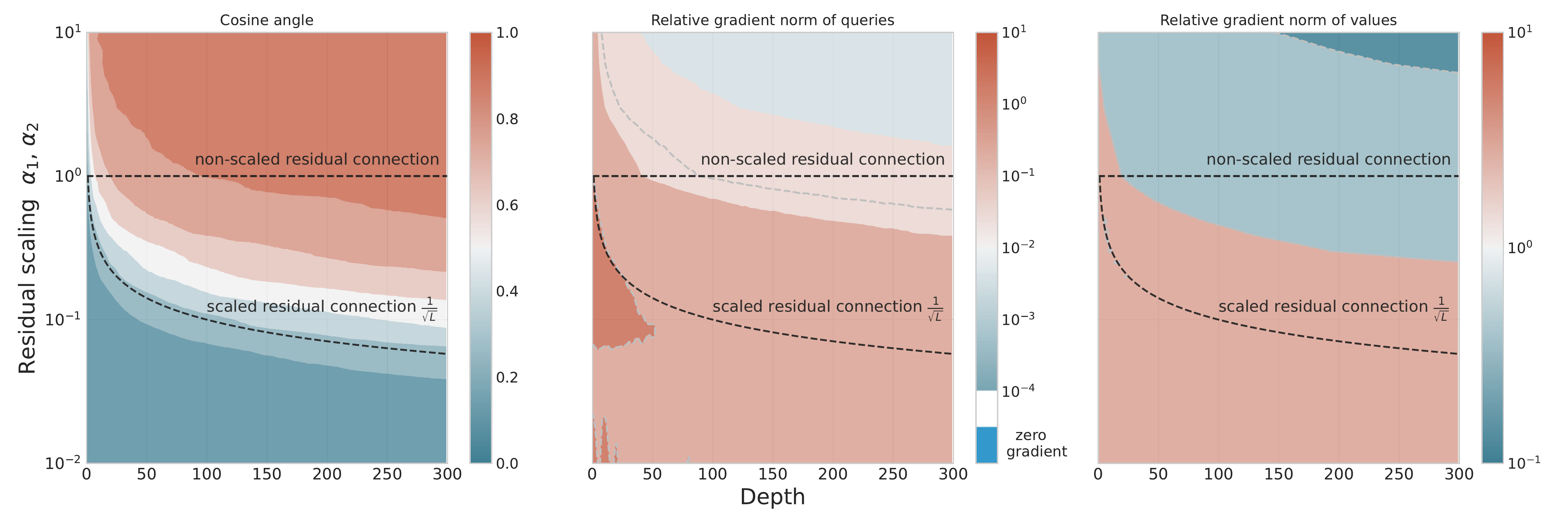}
    \caption{Same figure as~\ref{fig:residual_scaling} but with a PRE-LN architecture. The correlation at depth 0 originates from the correlations in the randomly initialized tokens' embeddings and positional encodings.}
    \label{fig:pre_ln_correlations}
\end{figure}

We present some additional results on the propagation of the norm and the correlations in Figure~\ref{fig:norm_corr_prop_1}. In particular, we empirically show that, with an adequate depth-dependent residual scaling, the norm and the correlation are stabilized, even for very deep networks. Furthermore, we demonstrate the propagation of the correlation and the gradient norms for the PRE-LN configuration in Figure~\ref{fig:pre_ln_correlations}. As also hinted in the main text, in Figure~\ref{fig:corr}, the increase in correlation with depth for PRE-LN is much less wild. This also results in better stabilized gradients for the queries and keys' parameters. We also observe the opposite trend for the gradients of the values, in relation to the POST-LN case in Figure~\ref{fig:residual_scaling}. We speculate that this different dependence, along with the better preserved correlation, is the main reason PRE-LN configured Transformers have been shown to scale better with depth. We plan to investigate this dependence more in future work.

\subsection{Further Empirical Assessment of Assumption \ref{ass:uniform_softmax}}
Here, we empirically test the accuracy and limitations of the uniform-attention assumption.

\begin{figure}
    \centering
    \includegraphics[scale=0.33]{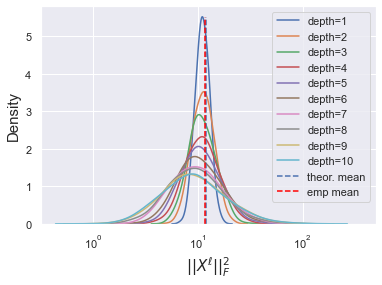}
    \includegraphics[scale=0.33]{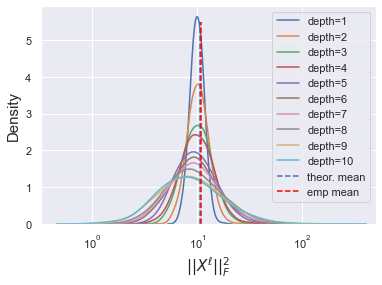}
    \includegraphics[scale=0.33]{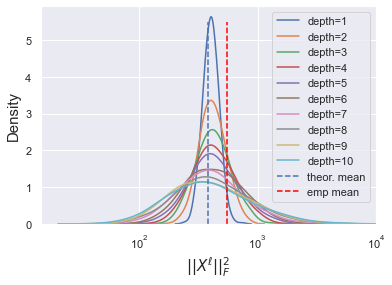}
    \caption{Density plots for $\norm{\Xm^\ell}_F^2$ for Transformers of depths $L$ from $1$ to $10$. The input $\Xm$ contains i.i.d Gaussian entries, simulating an embedding layer. We set $d:=d_v=d_q=30$. The empirical mean at $L=10$ is highlighted in a vertical dashed red line, while the theoretical mean (Lemma \ref{thm:forward_pass}) is a dashed blue line. The densities are estimated by sampling 1000 times the weights of the network. (Left): we adopt the uniform-attention. The standard deviation of the input is set to $1/\sqrt{d}$. (Center): Same, but removing the uniform-attention assumption. (Right): We remove the uniform-attention assumption, and set the standard deviation of the input to $1$.}
    \label{fig:norm_std_init_vs_uniform_att}
\end{figure}

For the empirical verification of Assumption \ref{ass:uniform_softmax} in the forward pass analysis, we plot the density of the norm of the representations for only-encoder Transformers of increasing depth. The results are shown in Fig \ref{fig:norm_std_init_vs_uniform_att}. Note that when the standard deviation of the input is set to $1/\sqrt{d}$, then the uniform-attention assumption provide an excellent approximation to the common Xavier-initialization. On the contrary, we observe a deviation when the standard deviation of the input is increased. Also, note how as the depth increases, the distribution becomes more heavy-tailed. This heavy-tailedness was recently formally shown for standard MLPs with and without ReLU activation \citep{noci2021precise, zavatone2021exact}.   

\begin{figure}[h]
    \centering
    \includegraphics[scale = 0.39]{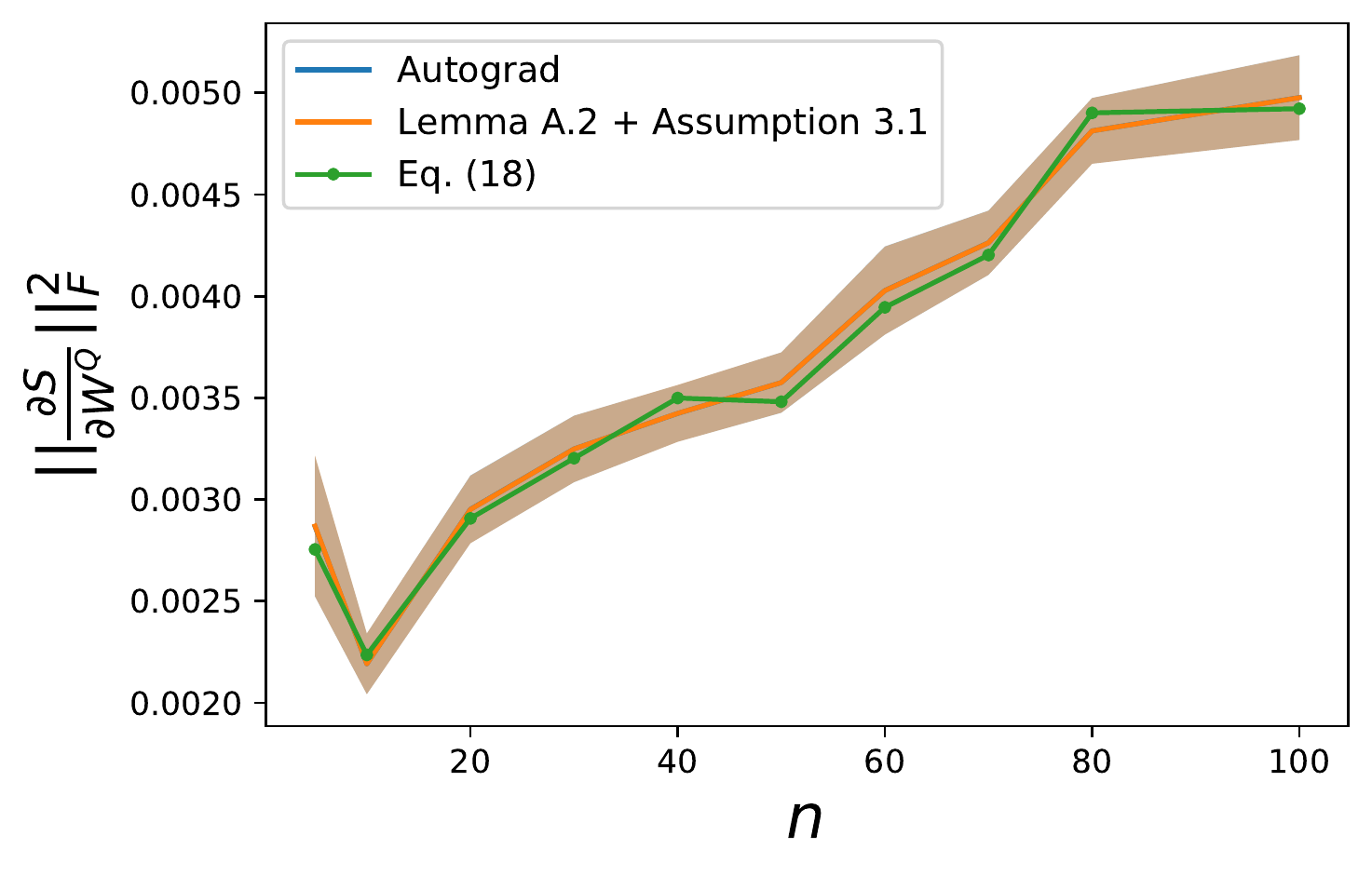}
    \includegraphics[scale = 0.39]{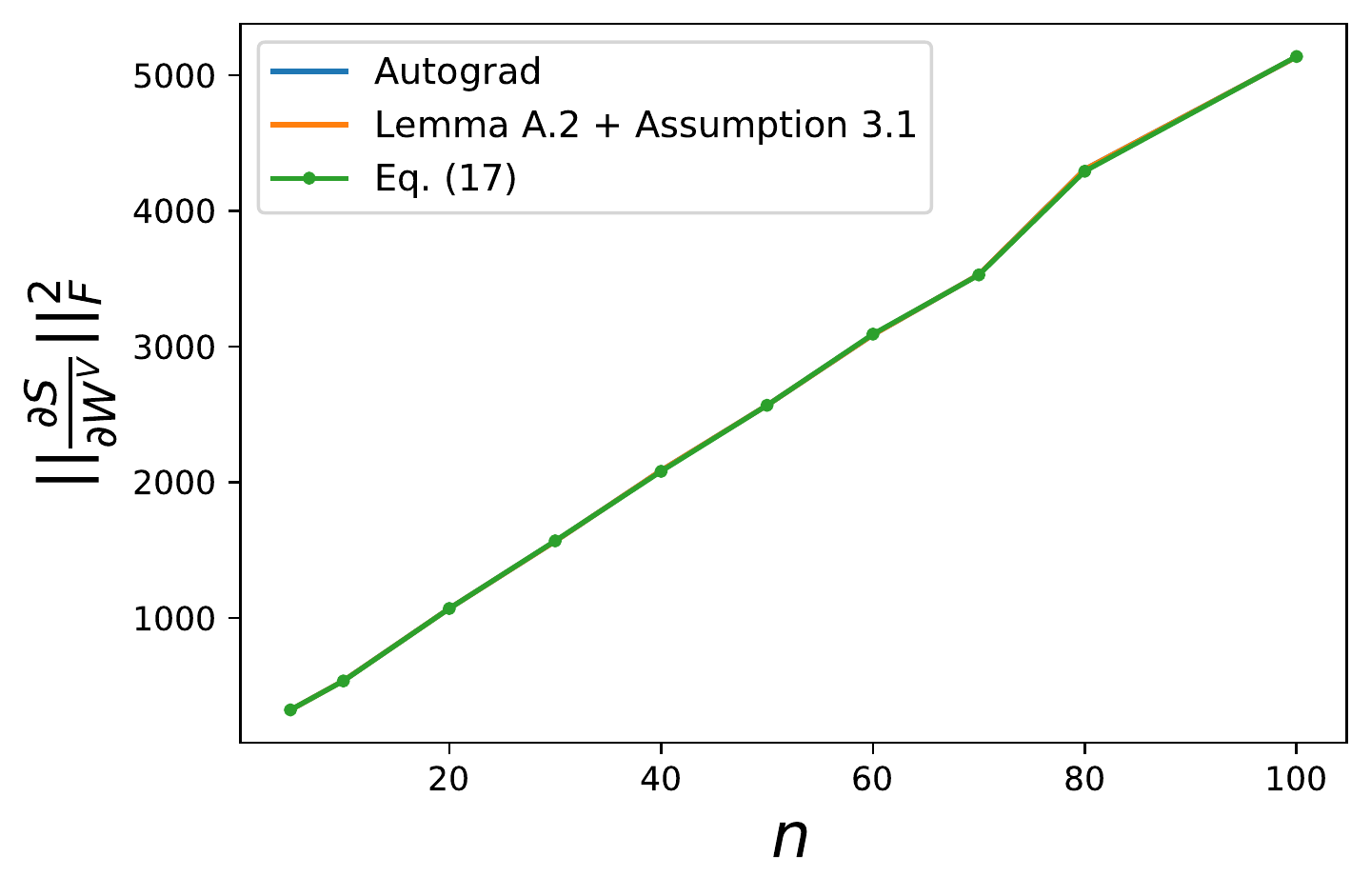}
    \includegraphics[scale = 0.39]{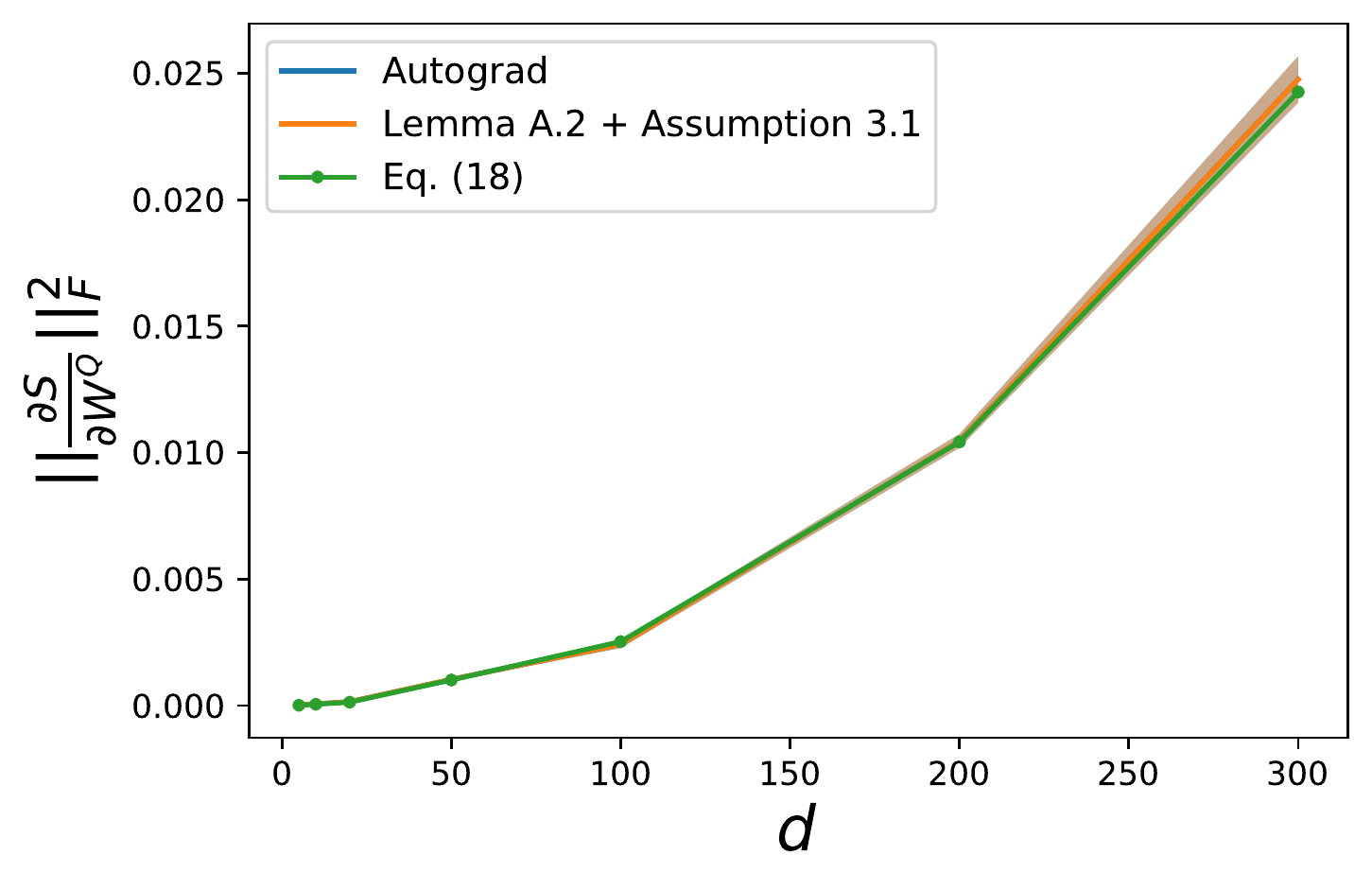}
    \includegraphics[scale = 0.39]{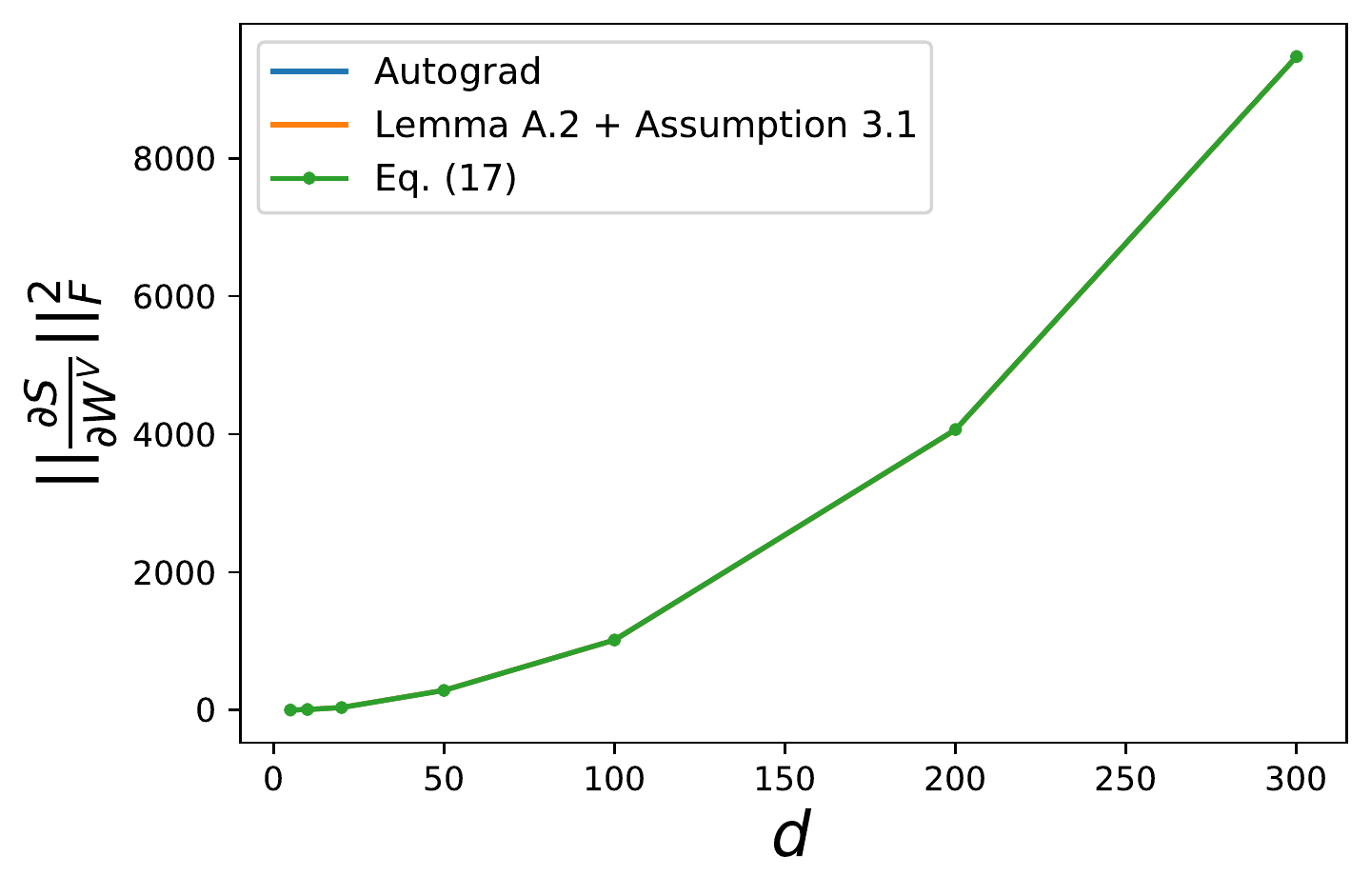}
    \includegraphics[scale = 0.39]{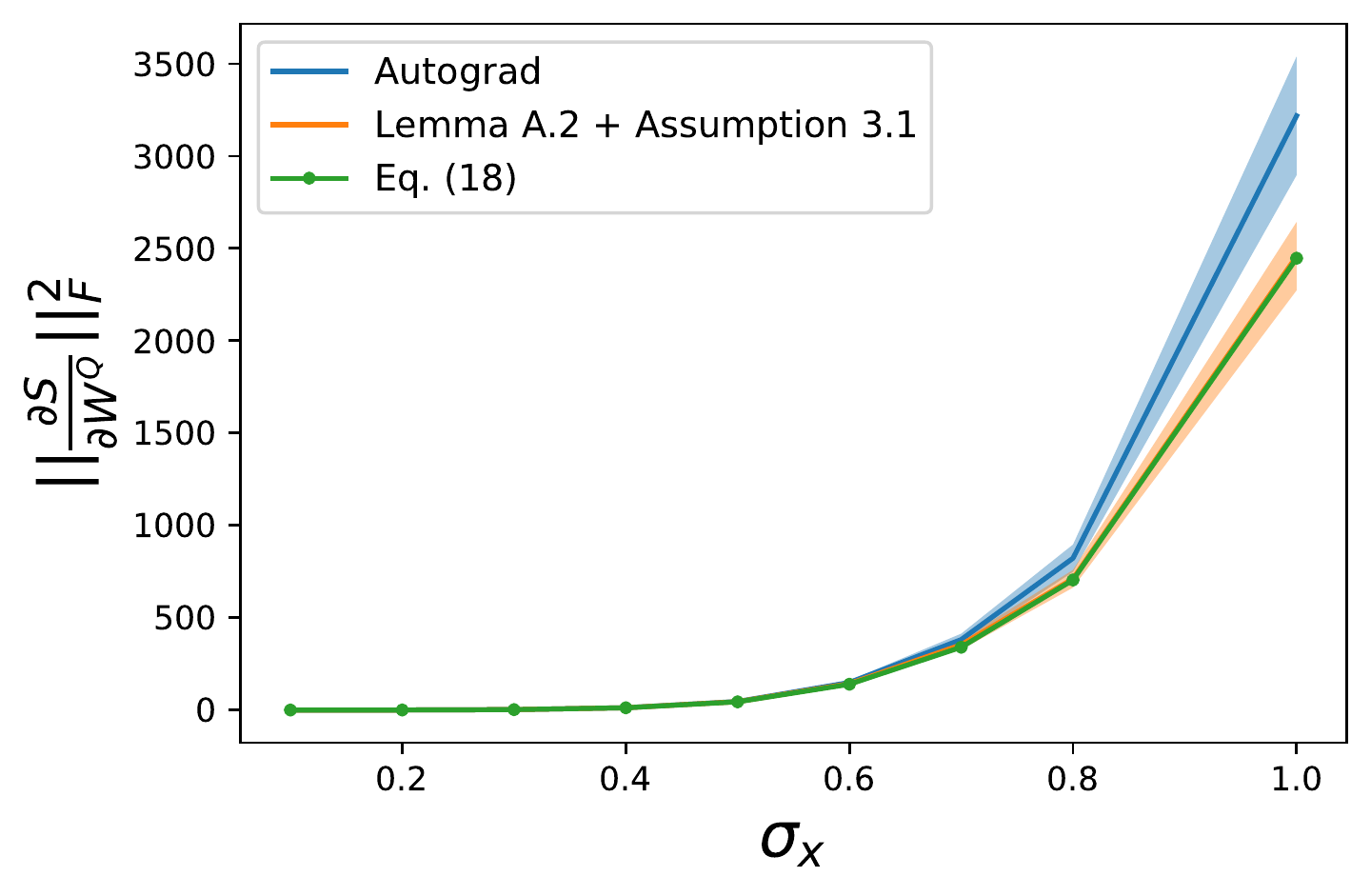}
    \includegraphics[scale = 0.39]{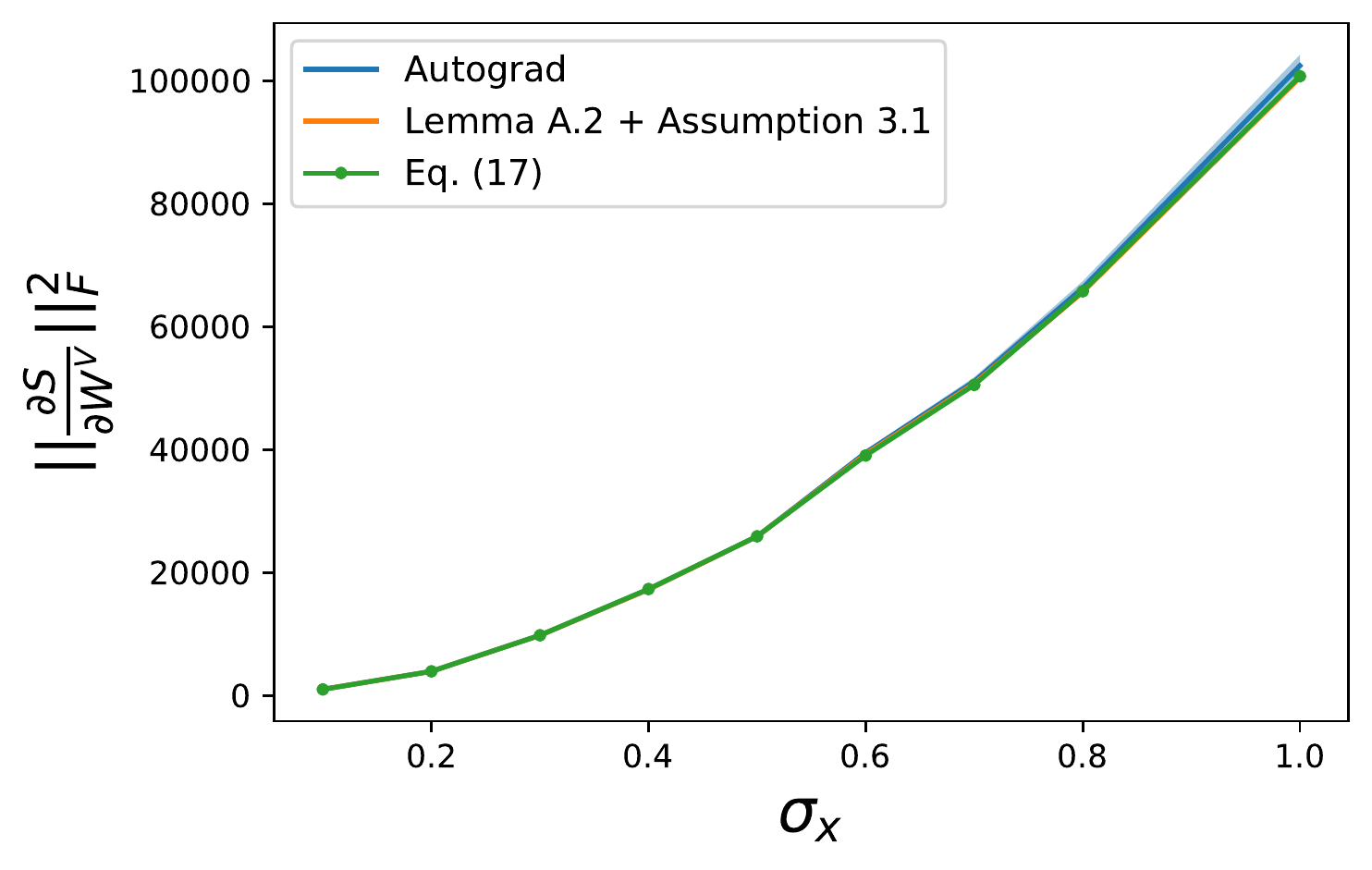}
    \includegraphics[scale = 0.39]{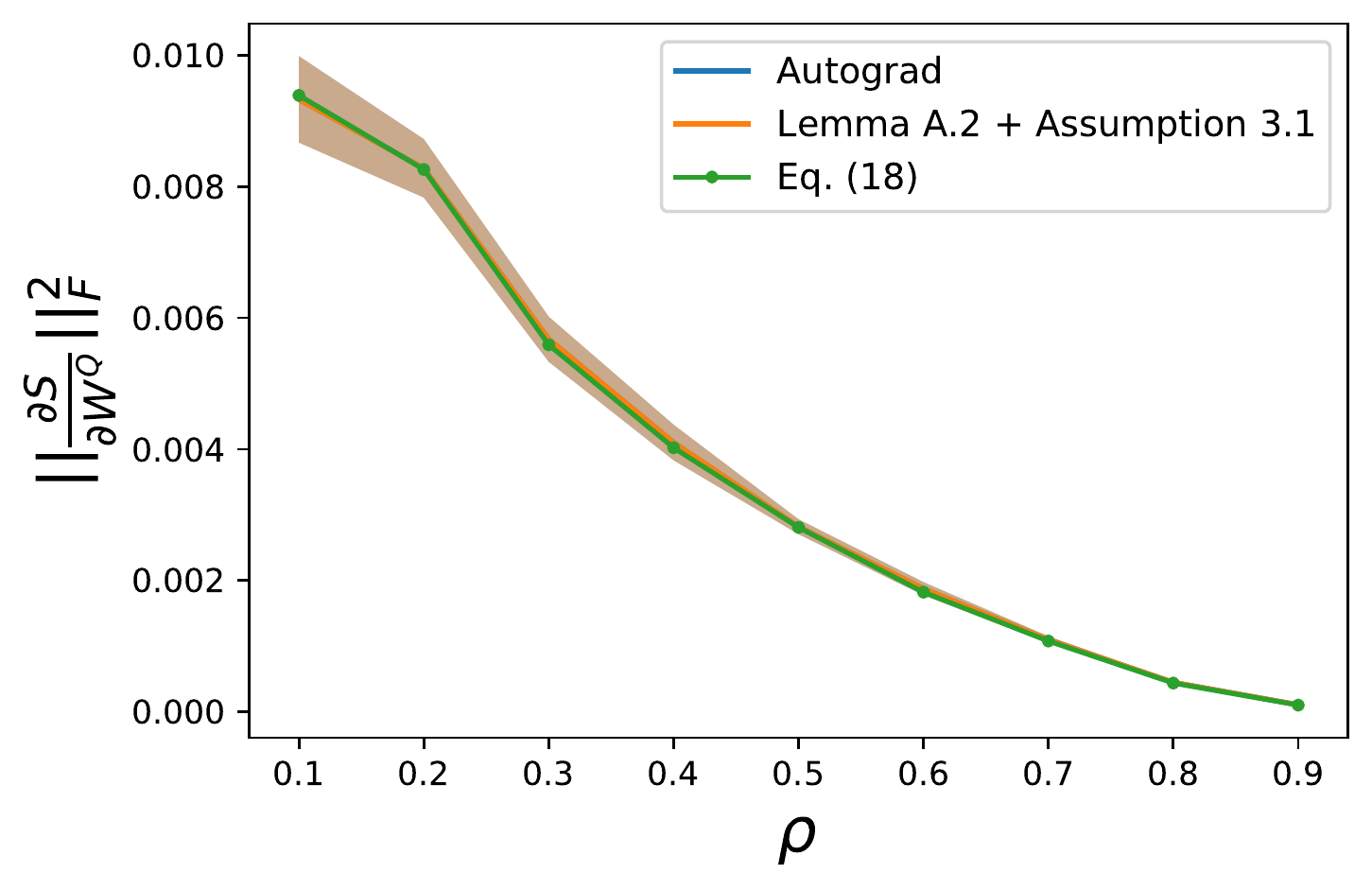}
    \includegraphics[scale = 0.39]{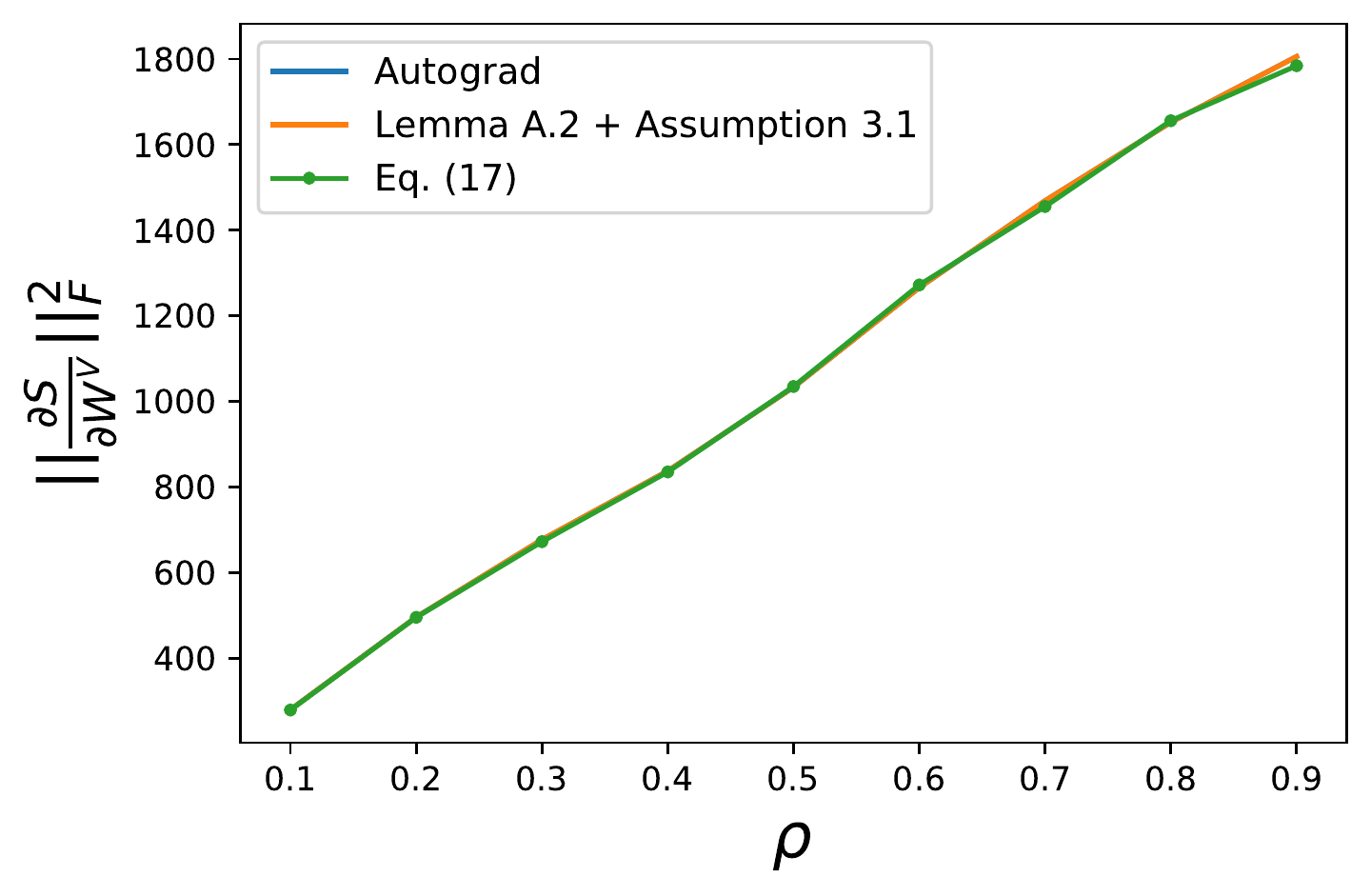}
    \caption{Empirical comparison of our theoretical findings. We sample, as aforementioned, the tokens according to a zero-mean Gaussian distribution, while varying the hidden dimension, sequence length, input correlation and input variance. Results are averaged over 20 runs.}
    \label{fig:empirical_verification}
\end{figure}

For the verification of the assumption in the backward pass, we additionally show in Fig. \ref{fig:empirical_verification} how the norm of the gradients w.r.t queries and keys depends on the hidden dimension, the sequence length, the input correlation and the input variance. \emph{Ground-truth} gradients are calculated with automatic differentiation, and they are compared with our theoretical results based on Assumption \ref{ass:uniform_softmax}. As shown in  Fig.\ref{fig:empirical_verification}, our theoretical predictions show a very good agreement with the true gradients. Again, we notice that the smaller the values of the input standard deviation the tighter the agreement of the theory with the simulations. Intuitively, a higher input variance causes the argument of the softmax to have a large range of values. This in turn causes a deviation from the uniform distribution (i.e. maximum entropy), towards the distribution of minimum entropy (a Delta Dirac, corresponding to attending to only one token).

\subsection{Empirical Verification of the Gradient Analysis of Section \ref{sec:dep_angle}}
\begin{figure}[t]
    \centering
    \includegraphics[width=1\linewidth]{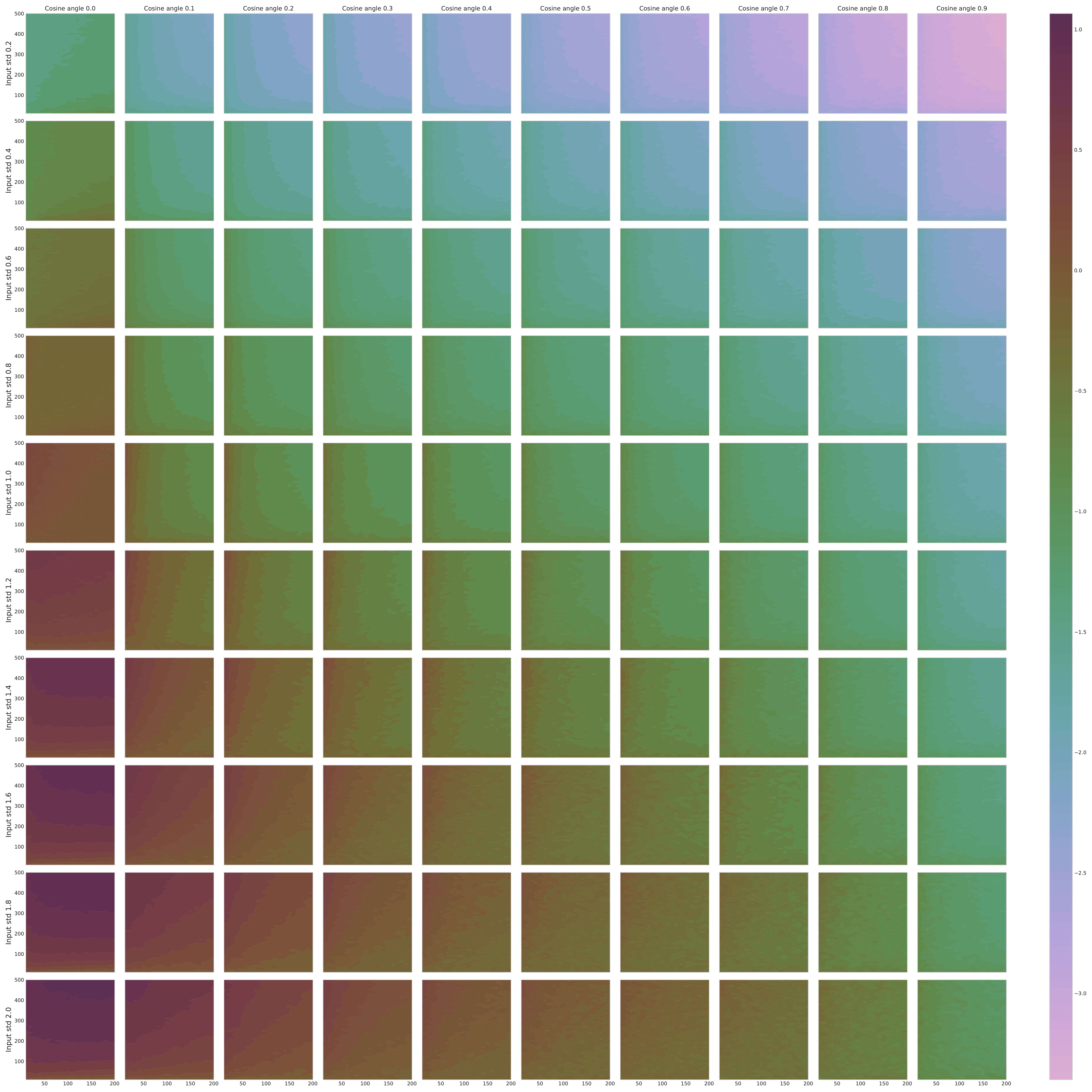}
    \caption{Log ratio of the norm of the gradients for the queries compared to those of the values for varying values of embedding dimension, sequence length, cosine of the tokens angle and standard deviation.}
    \label{fig:constant_correlation_factors}
\end{figure}

\begin{figure}[t]
    \centering
    \includegraphics[width=1\linewidth]{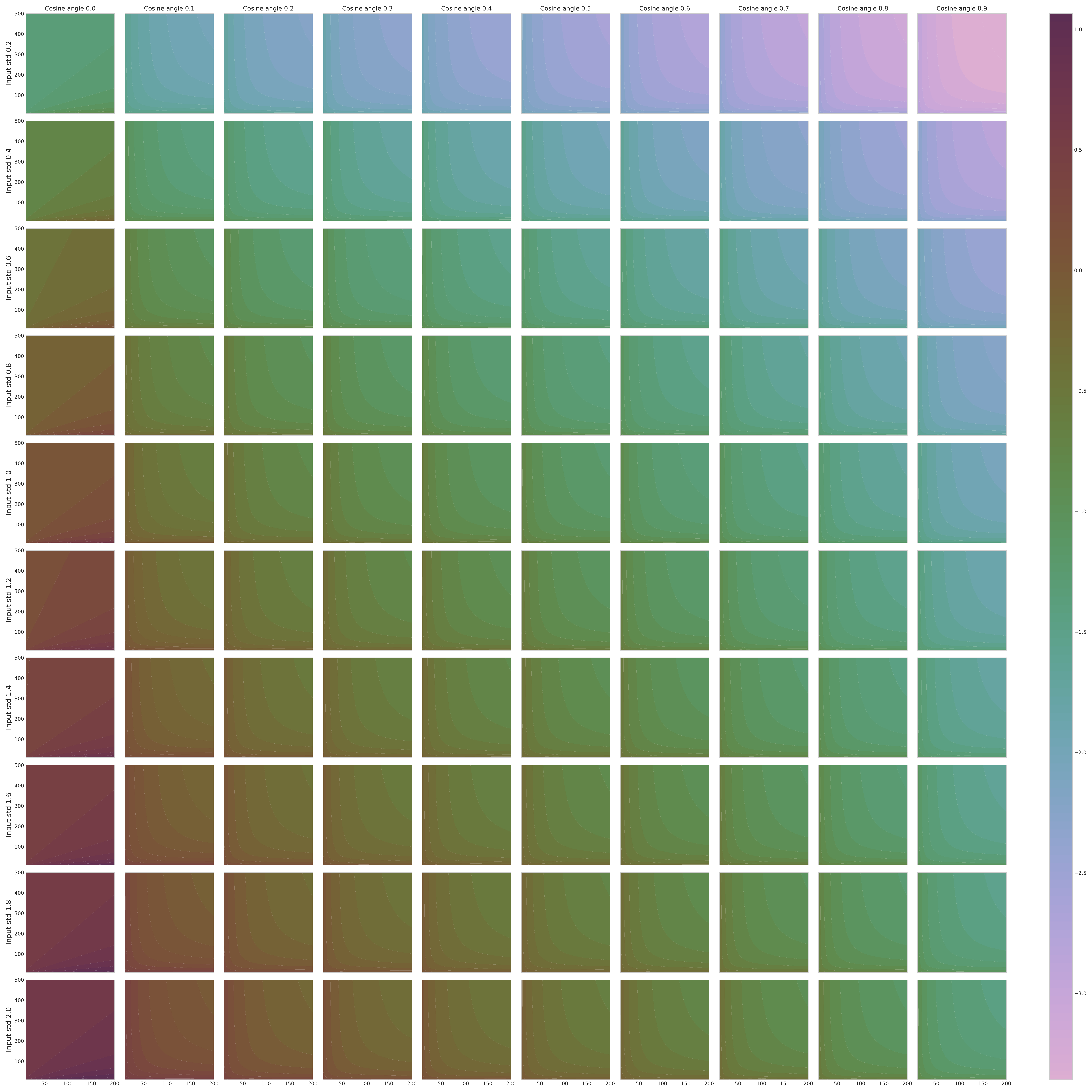}
    \caption{Log ratio of the norm of the gradients of the queries as expected by the theory, compared to those of the values for varying values of embedding dimension, sequence length, cosine of the tokens angle and standard deviation. We use Equations~\eqref{eq:grad_V} and~\eqref{eq:grad_Q}.}
    \label{fig:constant_correlation_factors_theory}
\end{figure}

Finally, in Figures~\ref{fig:constant_correlation_factors} and~\ref{fig:constant_correlation_factors_theory} we show the dependence of the norm of the gradients for the keys and values based on the parameters of the architecture and the task-specific parameters. Figure~\ref{fig:constant_correlation_factors} illustrates the true dependence and Figure~\ref{fig:constant_correlation_factors_theory} the one expected by the theory based on our assumptions. In short, the main takeaways are the following.
\begin{itemize}
    \item As the correlation between the tokens increases ($x$-axis in the global plot), the norm of the gradients of the queries quickly diminishes compared to the one of the values.
    \item The dependence on the variance of the input $\sigma_x^2$ is different ($y$-axis in the global plot), being linear for the values and cubic for the queries. This highlights the importance of a stabilized forward pass and provides another explanation regarding the successful use of layer norm in Transformers.
    \item The dependence on $n$ ($x$-axis in each subplot) and $d$ ($y$-axis in each subplot) is more complicated, also being a function of the correlation $\rho$ (compare the first column where $\rho = 0$ to the rest). 
\end{itemize}

\section{Experimental Setup}
\label{app:experimental_setup}

Here we provide more details regarding the experimental setup.

\subsection{Toy Example}

In Figure~\ref{fig:adam-adap}, we focus on a toy example where the task is to reverse a sequence of tokens. More specifically, given a sequence of $20$ numbers in the range $0-9$, we predict the same tokens in the inverted order. We use an embedding layer of size 16, initializes with variance 1, and sinusoidal positional encodings to initially embed the input. We use a 5-layer POST-LN Transformer encoder model, with a single head attention operation and a two-layer feed-forward layer with a ReLU nonlinearity. We use residual scaling in this case equal to $\alpha_1 = \alpha_2 = 1$. We train using Adam with betas parameters $(0.9, 0.999)$, learning rate $0.01$ and weight decay 0.

\subsection{Translation Task}

Introducing an inverse temperature scaling $\tau$ inside the softmax, modifies the attention operation to
\begin{equation}
    \Sm^{\ell} :=  \text{softmax}\left( \frac{\tau}{\sqrt{d_k}}\Xm^{\ell}\Wm^{Q}\left(\Xm^{\ell}\Wm^{K}\right)^\top \right) \Xm^\ell \Wm^{V}.
\end{equation}

\begin{figure}[t]
    \centering
    \includegraphics[width=1\linewidth]{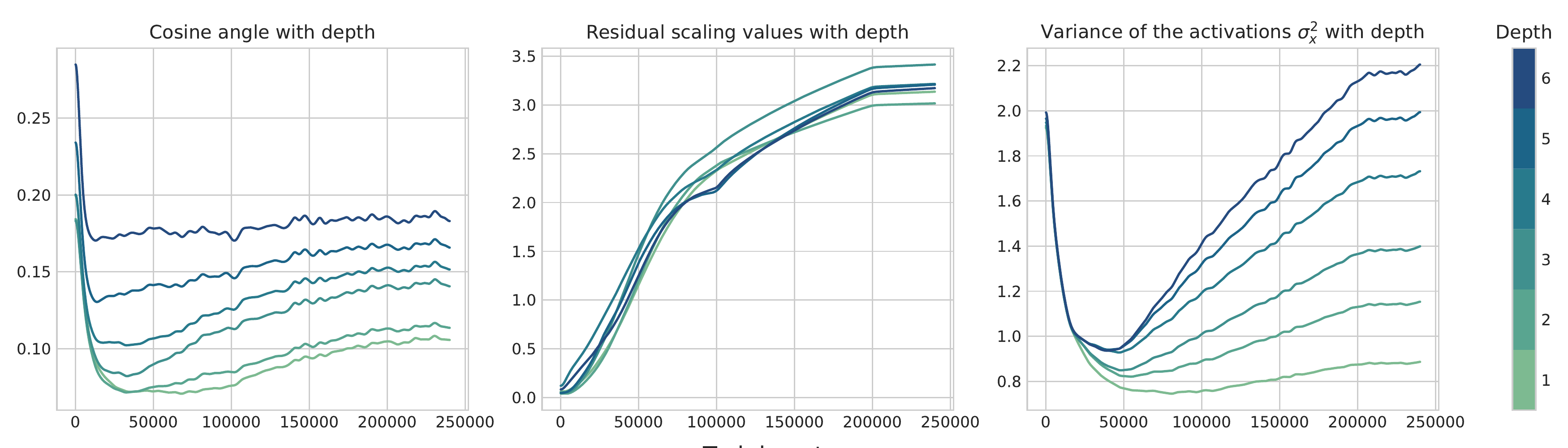}
    \caption{Evolution of the cosine of the angles, the trained residual $\alpha_1, \alpha_2$ and the activation norm throughout our training.}
    \label{fig:our_training}
\end{figure}

Then the gradient of the queries and keys parameters are directly scaled by this temperature value, following the same proof as for Eq.~\eqref{eq:jacobian_queries}. We choose a temperature value of $\tau_{\text{final}} = 8.5$ to match the gradient norms of the values and queries as in Equations.~\eqref{eq:grad_V} and~\eqref{eq:grad_Q}. Doing so, we assume a constant small correlation between tokens (also empirically verified in Fig.~\ref{fig:our_training}) and set the sequence length $n$ to the average found in our training dataset. Due to instabilities in training, we use warm-up on this temperature value. In short:
$$
\tau = \tau_{\text{final}} \cdot \text{max}(1, \frac{\text{step}}{\text{steps}_{\text{warmup}}}),
$$
with `$\text{steps}_{\text{warmup}}=1000$' and `step' the current training step.

We base our implementation on fairseq~\citep{ott2019fairseq}. For the hyperparameter configuration, we mostly rely on the extensive search already done in fairseq~\citep{ott2019fairseq} and~\cite{liu2020understanding}. The final used parameters are exhibited in Table~\ref{tab:hyperparameters}. For the final evaluation, we use the best-performing model on the left-out validation set. We apply weight decay as in~\cite{loshchilov2017decoupled} for both SGD and Adam.

 \begin{table}[h]
 \centering
 \begin{tabular}{cl|c}
 \hline
 & \multicolumn{1}{c|}{\textbf{Hyperparameters}} & \multicolumn{1}{c|}{\textbf{Value}}\\
 \hline
 \parbox[t]{2mm}{\multirow{9}{*}{\rotatebox[origin=c]{90}{}}} & Max tokens & 4096 \\
  & Label smoothing & 0.1 \\
  & clip-norm & 0.0 \\
  & General Dropout & 0.3 \\
  & Attention Dropout & 0.1 \\
  & ReLU Dropout & 0.1 \\
  & Hidden size & 512 \\
  & FFN inner hidden size & 2048 \\
  & Attention Heads & 4 \\
  \hline
  \parbox[t]{2mm}{\multirow{7}{*}{\rotatebox[origin=c]{90}{Adam}}} & Learning rate & $7\epsilon^{-4}$ \\
  & Learning rate scheduler & inverse sqrt \\
  & Warm-up updates & 6000 \\
  & Warm-up init learning rate & 1e-7 \\
  & Adam $(\beta_1, \beta_2)$ & (0.9, 0.98) \\
  & Training updates & 100K \\
  & Weight decay & 0.0001 \\
  \hline
  \parbox[t]{2mm}{\multirow{6}{*}{\rotatebox[origin=c]{90}{SGD}}} & Learning rate & $2\epsilon^{-2}$ \\
  & Learning rate scheduler & step \\
  & Step scheduler $\gamma$ & 0.1 \\
  & Step scheduler update steps & [100K, 200K] \\
  & Training updates & 250K \\
  & Weight decay & 0.001 \\
 \hline
\end{tabular}
\caption{Hyperparameters for the IWSLT'14 De-En translation task.}
\label{tab:hyperparameters}
\end{table}

Finally, in Figure~\ref{fig:our_training} we display the evolution of correlations, residual scaling, and norm of the activations, with depth, for our best trained model. The residual scaling $\alpha_1, \alpha_2$ are trainable parameters. This enables them to weight differently the residual branches if deemed necessary. Although these values increase during training, the correlation between the tokens does not significantly increase, which as implied by our main results, allows efficient propagation of the gradients. The norm of the propagated forward signal tends to slightly increase with depth.

\end{document}